%% file: acl_latex.tex
\documentclass[11pt]{article}

\usepackage[final]{acl}

\usepackage{times}
\usepackage{latexsym}

\usepackage[T1]{fontenc}

\usepackage[utf8]{inputenc}

\usepackage{microtype}

\usepackage{graphicx}
\usepackage{multirow}
\usepackage{xurl}
\usepackage{amsmath}
\usepackage{amssymb}
\usepackage{booktabs}
\usepackage{tabularx}
\usepackage{tikz}
\usepackage{colortbl}
\usepackage{comment}
\usepackage{pgfplots}
\pgfplotsset{compat=1.18}
\usepgfplotslibrary{groupplots}
\usepackage{tikz, pgfplots}
\usetikzlibrary{pgfplots.groupplots}

\definecolor{GreenLight}{HTML}{C3E6CB} 
\definecolor{GreenDark}{HTML}{C3E6CB}  

\definecolor{RedLight}{HTML}{FCE8E6}   
\definecolor{RedDark}{HTML}{FAD2CF}    

\definecolor{brickred}{rgb}{0.8, 0.25, 0.33}
\definecolor{applegreen}{rgb}{0.55, 0.71, 0.0}
\definecolor{asparagus}{rgb}{0.53, 0.66, 0.42}
\definecolor{darkpastelgreen}{rgb}{0.01, 0.75, 0.24}

\newcommand{\posL}{\cellcolor{darkpastelgreen!50}}  
\newcommand{\posS}{\cellcolor{darkpastelgreen!20}} 

\newcommand{\negS}{\cellcolor{brickred!20}}   
\newcommand{\negL}{\cellcolor{brickred!50}}    

\renewcommand{\texttt}[1]{\begingroup\ttfamily\hyphenchar\font=`\-#1\endgroup}
\usetikzlibrary{shapes.geometric, arrows.meta, positioning, fit, backgrounds, shadows, calc, patterns}
\usepackage{hyperref}

\title{An Empirical Study on Preference Tuning Generalization and Diversity Under Domain Shift}

\author{
Constantinos Karouzos \qquad
Xingwei Tan \qquad
Nikolaos Aletras\\
School of Computer Science \\
University of Sheffield, UK \\
\texttt{\{kkarouzos1, xingwei.tan, n.aletras\}@sheffield.ac.uk}
}

\begin{document}
\maketitle
\begin{abstract}
Preference tuning aligns base language models to human judgments of quality, helpfulness, or safety by optimizing over explicit preference signals rather than likelihood alone. Prior work has shown that preference tuning degrades performance and reduces helpfulness outside the training domain. However, the extent to which adaptation strategies mitigate this domain shift remains unexplored. We address this challenge by conducting a comprehensive and systematic study of alignment generalization under domain shift. We compare five popular alignment objectives and various adaptation strategies from source to target, including target-domain supervised fine-tuning and pseudo-labeling, across summarization, question-answering helpfulness, and safety alignment tasks. Our findings reveal systematic differences in generalization across alignment objectives under domain shift. We show that adaptation strategies based on pseudo-labeling substantially reduce domain-shift degradation but induce mode collapse, revealing a generalization–diversity trade-off.\footnote{Code available at: \url{https://github.com/ckarouzos/prefadap}}
\end{abstract}

\section{Introduction}
\label{sec:intro}

Large language models (LLMs), such as GPT-5 \citep{openai2025gpt5}, Gemini 3 \citep{Google025Gemini3}, and DeepSeek-V3 \citep{deepseekai2025deepseekv3technicalreport}, rely on post-training, i.e., human preference optimization beyond pretraining to improve helpfulness, safety, and truthfulness \citep{ouyang2022training}. Post-training, consisting of supervised fine-tuning (SFT) and preference-based optimization, is a standard component of LLM development \citep{lambert2025reinforcement}.

Despite widespread adoption, existing work has not systematically characterized the generalization of preference optimization methods under domain shift. Prior work provides limited evidence of out-of-domain generalization, typically for a single objective in isolation. For example, it focuses only on either Direct Preference Optimization \citep[DPO]{rafailov2023direct}, or as an analysis tool \citep{kirk2023understanding} of reinforcement learning from human feedback~\citep[RLHF]{ouyang2022training} with Proximal Policy Optimization \citep[PPO]{schulman2017proximal}. Moreover, no systematic evaluation compares other preference objectives or analyzes how adaptation strategies mitigate domain shift. 

\input{latex/figure_final.tex}

We address this gap via a study along two practical axes. The first is the choice of the alignment objective, covering a broad spectrum of paradigms: from standard SFT and online reinforcement learning, RLHF and group relative policy optimization \citep[GRPO]{shao2024deepseekmath}, to offline, RL-free formulations including DPO \citep{rafailov2023direct}, Kahneman--Tversky Optimization \citep[KTO]{ethayarajh2024kto}, and odds-ratio preference optimization \citep[ORPO]{hong-etal-2024-orpo}. The second axis is the choice of domain adaptation strategy, ranging from target-domain SFT to pseudo-labeling, with model generated preferences.

We test alignment objectives and domain adaptation methods across three complementary testbeds, each consisting of a source and target domain. The first is a summarization task adapting from informal Reddit \textit{TL;DR} posts \citep{volske-etal-2017-tl} to formal CNN/DailyMail (\textit{CNN/DM}) news articles \citep{nallapati-etal-2016-abstractive}. The second is a near domain helpfulness focused question answering task transferring from \textit{AskEngineers} to \textit{AskCulinary} of the Stanford Human Preferences (SHP) dataset \citep{pmlr-v162-ethayarajh22a}. The third is a safety alignment transfer between harm categories in PKU-SafeRLHF \citep{ji-etal-2025-pku}, from Cybercrime to Violence and Physical Harm. 
Figure~\ref{fig:framework} summarizes the experimental framework. Our contributions:
\begin{itemize}
\setlength\itemsep{0em}
\item A comparison of five preference optimization objectives, alongside SFT, under domain shift, across summarization, helpfulness, and safety.
\item Evidence that adaptation strategy, especially pseudo-labeling, often matters more than the alignment objective.
\item An analysis of the resulting generalization--diversity trade-off, including mode collapse under synthetic target supervision.
\end{itemize}

\section{Related Work}
\label{sec:related_work}

\paragraph{Preference Alignment.}
Alignment has evolved from SFT to RLHF, where a reward model (RM) guides policy updates \citep{christiano2017deep, stiennon2020learning, ouyang2022training}. Recent work analyzes RLHF as divergence estimation \citep{chaudhari2025rlhfdeciphered, haldar2025llm}, which often suffers from training instability  \citep{rafailov2023direct}. This has motivated DPO and other RL-free variants to improve stability \citep{meng2024simpo, ethayarajh2024kto, zhao2023slic, cho-etal-2025-rethinking, wangvpo, guo2025proximalized}. Concurrently, reference-free and odds-ratio methods  integrate alignment directly into language modeling or multi-objective frameworks \citep{hong-etal-2024-orpo, bansal-etal-2025-comparing, liu-etal-2025-amopo, chenmallowspo, luo-etal-2025-rethinking}. Further extensions view alignment through a game-theoretic or group-based lens, such as GRPO and Nash-style self-play \citep{yao2025group, zhu-etal-2025-reinforcement, zhu2025gtalign, wu2024self, tang2025game, zhou2025extragradient}.

\paragraph{Domain Adaptation Strategies.}
Standard adaptation relies on domain-adaptive pretraining, i.e., continuing the pretraining phase on domain-specific data \citep{gururangan-etal-2020-dont, kirkpatrick2017overcoming}. Other approaches leverage synthetic supervision via AI teachers (reinforcement learning from AI feedback; RLAIF) or self-play \citep{bai2022constitutional, lee2023rlaif, chen2024selfplay, wang-etal-2023-self-instruct}. Work on preference data construction highlights the importance of filtering hard negatives and synthesizing high-quality pairs \citep{he2025air, xiao-etal-2025-finding}, alongside data selection curricula that match difficulty to model competence \citep{deng2025less, miranda-etal-2025-hybrid, zhang-etal-2025-preference}. Complementary approaches model distribution shifts directly through robust preference estimation, multi-supervisor reweighting, and weak-to-strong generalization \citep{huang2025distribution, yan-etal-2025-rpo, geng2025delta, zhu2024weak, belakaria2025sharpe, patel2025adapt}. 

\paragraph{Alignment Robustness and Diversity.}
Optimizing for safety or helpfulness often incurs an \textit{alignment tax} on out-of-domain performance \citep{lin-etal-2024-mitigating, balepur-etal-2025-good}. In summarization, this appears as poor transfer between topics \citep{kornilova-eidelman-2019-billsum, delucia2025can, afzal-etal-2024-adapteval}. These failures stem from mode collapse and reduced variability. Recent work proposes diversity-aware objectives to mitigate typicality bias \citep{zhang2025verbalized, guo2025benchmarking, cao-etal-2025-perspective, lanchantin2025diverse, ismayilzada-etal-2025-creative}. While distributionally robust optimization and pluralistic alignment aim to preserve diverse behavior \citep{xu2025robust, golz2025distortion, lake-etal-2025-distributional, yao2025both}, empirical comparisons of how standard objectives affect diversity under domain shift remain limited.

\section{Methodology}
\label{sec:method}

\subsection{Problem Setting}
\label{subsec:problem_setting}

We study domain adaptation for aligning models to human preferences when annotations are unavailable. We train a policy, $\pi_\theta$, to generate outputs $y \in \mathcal{Y}$ for prompts $x \in \mathcal{X}$ in a source domain and evaluate on a target domain. The \textbf{source domain} ($\mathcal{D}_S$) consists of a labeled preference dataset $\mathcal{D}^{\text{pref}}_S$. The format varies by objective: prompt-demonstration pairs $(x_i, y_i^*)$; preference triplets $(x_i, y_i^w, y_i^l)$ where $y_i^w \succ y_i^l$; or binary-labeled examples $(x_i, y_i, l_i)$ with $l_i \in \{\text{desirable}, \text{undesirable}\}$. The \textbf{target domain} $\mathcal{D}_T$, consists of prompts $\{x_j\}_{j=1}^M$ and responses $y_j$ without preference annotations. The key challenge is the distributional shift between $\mathcal{D}_S$ and $\mathcal{D}_T$, involving style, topic, or implicit preference criteria. We use the source preference data $\mathcal{D}^{\text{pref}}_S$ and the target-domain corpus to learn a policy $\pi_\theta$ that generalizes to $\mathcal{D}_T$, producing high quality outputs, where quality reflects helpfulness, stylistic appropriateness, or safety.

\subsection{Preference Optimization Objectives}
\label{subsec:optimization_objectives}

\paragraph{DPO.} It directly optimizes the policy from preference pairs, using a Bradley-Terry objective. Let $\pi_{\mathrm{ref}}$ denote a fixed reference policy:
\begin{equation*}
\label{eq:dpo}
\displaystyle
\mathcal{L}_{\text{DPO}}(\pi_\theta; \pi_{\mathrm{ref}}) = -\mathbb{E}_{(x, y_w, y_l) \sim \mathcal{D}_S} \left[ \log \sigma(\Delta) \right],
\end{equation*}

\noindent where $\Delta = \beta \log \frac{\pi_\theta(y_w|x)}{\pi_{\mathrm{ref}}(y_w|x)} - \beta \log \frac{\pi_\theta(y_l|x)}{\pi_{\mathrm{ref}}(y_l|x)}$, and $\beta$ is a temperature parameter.

\paragraph{KTO.} KTO uses binary feedback. The loss encourages higher likelihoods for desirable examples and lower likelihoods for undesirable ones. The full loss is an expectation over per-example terms:
\begin{equation*}
\label{eq:kto_main}
\begin{aligned}
\mathcal{L}_{\text{KTO}}(\pi_\theta;\pi_{\mathrm{ref}})
&=\mathbb{E}_{(x,y,l)\sim\mathcal{D}_S}
\left[\mathcal{L}_{\text{term}}(x,y,l)\right].
\end{aligned}
\end{equation*}

\noindent where the loss term $\mathcal{L}_{\text{term}}$ depends on the label $l$:
\begin{equation*}
\label{eq:kto_term}
\mathcal{L}_{\text{term}}=
\begin{cases}
\lambda_D[1-\sigma(s_\theta)], & l=D,\\
\lambda_U[1-\sigma(-s_\theta)], & l=U,
\end{cases}
\end{equation*}
\begin{equation*}
\begin{aligned}
r_\theta(x,y)
&=\log\frac{\pi_\theta(y\mid x)}
{\pi_{\mathrm{ref}}(y\mid x)},\\
d_\theta(x')
&=D_{\mathrm{KL}}\!\left(
\pi_\theta(\cdot\mid x')\,\|\,
\pi_{\mathrm{ref}}(\cdot\mid x')\right),\\
z_0&=\mathbb{E}_{x'\sim\mathcal{D}_S}[d_\theta(x')],\\
s_\theta&=\beta\left(r_\theta(x,y)-z_0\right).
\end{aligned}
\end{equation*}

\noindent $D$ and $U$ are desirable and undesirable labels, $r_\theta$ is the implicit reward, $d_\theta(x')$ is the prompt-level KL divergence, and $z_0$ is the KL reference point. The coefficients $\lambda_D$ and $\lambda_U$ weight the two classes.

\paragraph{ORPO.} A single-stage, reference-free method that combines a standard language modeling loss on the winning response with an odds-ratio preference term.
\begin{equation*}
\begin{aligned}
\mathcal{L}_{\mathrm{ORPO}}(\theta)
&=\mathbb{E}_{(x,y_w,y_l)\sim\mathcal{D}_S}\Big[
-\log p_\theta(y_w\mid x)\\
&\qquad-\lambda\log\sigma\!\left(
\log\frac{o_\theta(y_w\mid x)}
{o_\theta(y_l\mid x)}\right)\Big].
\end{aligned}
\end{equation*}

\paragraph{PPO.}
We apply RLHF with PPO \citep{schulman2017proximal} in two stages. First, we train a RM $r_\phi(x,y)$ to minimize the pairwise ranking loss:
\begin{equation*}
\begin{aligned}
\mathcal{L}_{\text{RM}}(\phi)
&= 
-\mathbb{E}_{(x, y_w, y_l) \sim \mathcal{D}_S} 
\Big[
\log \sigma \big(
r_\phi(x, y_w) 
\\
&\qquad\qquad
- r_\phi(x, y_l)
\big)
\Big].
\end{aligned}
\end{equation*}
\noindent Then, we optimize the policy $\pi_\theta$ to maximize the expected reward while penalizing deviation from the reference model $\pi_{\text{ref}}$ via KL-divergence:
\begin{equation*}
\mathcal{L}_{\text{PPO}}(\theta) = \mathbb{E}_{x, y \sim \pi_\theta} \left[r_\phi(x, y) - \beta \log \frac{\pi_\theta(y|x)}{\pi_{\text{ref}}(y|x)}\right].
\end{equation*}

\paragraph{GRPO.} GRPO samples a group of outputs per prompt $x$ and computes an advantage $A_i$ for each output relative to the group average:
\begin{equation*}
A_i = \frac{r_i - \text{mean}(\{r_1, \dots, r_G\})}{\text{std}(\{r_1, \dots, r_G\}) + \eta}.
\end{equation*}

\noindent We maximize the surrogate objective, similar to PPO but without a value network:
\begin{equation*}
\begin{aligned}
&\mathcal{J}_{\text{GRPO}}(\theta)
=
\mathbb{E}
\Bigg[
\frac{1}{G}\sum_{i=1}^{G}\frac{1}{|y_i|}
\sum_{t=1}^{|y_i|}
\Big(
\min\big(
\rho_{i,t} A_i,
\\[-0.2em]
&\quad
\operatorname{clip}(\rho_{i,t}, 1-\epsilon, 1+\epsilon) A_i \big)
-
\beta D_{\mathrm{KL}}^{i,t}
\Big)
\Bigg],
\end{aligned}
\end{equation*}

\noindent where $\rho_{i,t}=\frac{\pi_\theta(y_{i,t}|x,y_{i,<t})}{\pi_{\theta_{\mathrm{old}}}(y_{i,t}|x,y_{i,<t})}$ is the token-level probability ratio, and $D_{\mathrm{KL}}^{i,t}$ is the KL estimator.

\subsection{Domain Adaptation Strategies}
\label{subsec:adaptation_strategies}

\paragraph{SFT.}
We use SFT to adapt policies by minimizing the negative log-likelihood of $y$ given $x$. The training data is drawn from one of four configurations: the \textit{source domain} ($\mathcal{D}_S$), the \textit{target domain} ($\mathcal{D}_T$), a \textit{mixture of both} ($\mathcal{D}_{S+T}$), or the \textit{target domain via pseudo-labeling} ($\mathcal{D}_T^{\text{synth}}$). 

\paragraph{Pseudo-Labeling.}
We create a synthetic preference dataset for the target domain. We distill the preference priors of a larger \textit{teacher} model into in-domain training signals for the student. For each prompt $x$ in the unlabeled target domain corpus $\mathcal{D}_T$, we generate multiple candidate responses $\{y_1, ..., y_k\}$. We construct preference pairs $(x, y_w, y_l)$ by designating the teacher-generated candidate as the preferred response $y_w$ and the original reference response from the dataset as the dispreferred response $y_l$. The resulting $\mathcal{D}_T^{\text{synth}}$ is formatted per objective: SFT uses the prompt and $y_w$; DPO, ORPO and RMs use the generated pairs; KTO unpairs them into binary labeled examples. For online RL, we train a reward model on $\mathcal{D}_T^{\text{synth}}$ and optimize the policy on the target-domain. 

\section{Experimental Setup}
\label{sec:experimental}

\subsection{Testbeds and Data}
\label{subsec:datasets}

\paragraph{Summarization: Reddit TL;DR $\rightarrow$ CNN/DM.}
The source domain ($\mathcal{D}_S$) consists of informal Reddit TL;DR summaries \citep{volske-etal-2017-tl}, and the target ($\mathcal{D}_T$) of formal CNN/DM news highlights \citep{nallapati-etal-2016-abstractive}. We use the human preference annotations ($\mathcal{D}^{\text{pref}}_S$) of \citet{stiennon2020learning}.

\paragraph{QA Helpfulness: AskEngineers $\rightarrow$ AskCulinary.}
We use the question answering SHP dataset \citep{pmlr-v162-ethayarajh22a}, assigning \textit{AskEngineers} as $\mathcal{D}_S$ and \textit{AskCulinary} as $\mathcal{D}_T$. Each example has a prompt and a pair of responses with a human preference. This is a near domain shift: the surface format is constant (Q\&A platform, Reddit), but the epistemic style and pragmatic domain change.

\paragraph{Safety: Cybercrime $\rightarrow$ Violence/Physical~Harm.} We use PKU-SafeRLHF \citep{ji-etal-2025-pku}, filtering into non-overlapping harm categories. The $\mathcal{D}_S$ contains Cybercrime examples and $\mathcal{D}_T$ contains Violence and Physical Harm examples. Each $\mathcal{D}_S$ example includes a prompt with a pair of responses annotated for relative safety.

\subsection{Base Models}
\label{subsec:base_models}

We train two open weight models, \texttt{Llama-3.1-8B} \citep{grattafiori2024llama3herdmodels} and \texttt{OLMo-3-7B} \citep{olmo2025olmo3}. We start from base pretrained models to isolate the effect of each post-training stage.

\subsection{Training Settings}
\label{subsec:baselines}
We define training settings by the data used in SFT and preference optimization (Pref.) stages: \textbf{Base}, the pre-trained model before alignment; \textbf{SFT baselines} on $\mathcal{D}_S$, $\mathcal{D}_T$, $\mathcal{D}_{S+T}$, or $\mathcal{D}_T^{\text{synth}}$ without preference tuning; \textbf{Source-only}, the standard two-stage process on source data ($\mathcal{D}_S \to \mathcal{D}_S$); \textbf{Direct alignment}, preferences on $\mathcal{D}_S$ without SFT, for DPO/KTO/ORPO; \textbf{Mix-SFT}, the SFT stage uses a mixture of source and target data ($\mathcal{D}_{S+T} \to \mathcal{D}_S$); \textbf{Target-SFT}, the SFT stage relies on target-domain data, followed by preference optimization on the source data ($\mathcal{D}_T \to \mathcal{D}_S$); and \textbf{Pseudo-labeled}, where both stages use the synthetic target-domain data ($\mathcal{D}_T^{\text{synth}} \to \mathcal{D}_T^{\text{synth}}$).

\subsection{Evaluation}
\label{subsec:evaluation_metrics}

\paragraph{LLM-as-a-judge win rate.}
Following \citet{rafailov2023direct} and \citet{kirk2023understanding}, we use an LLM-as-a-judge. For each prompt, we compare the adapted model output and a reference response: the \texttt{reference} summary for summarization; the \texttt{chosen} response for QA Helpfulness. The LLM-as-a-judge selects which response better satisfies task-specific criteria. Appendix~\ref{ref:appendix_prompts} provides the judge prompt templates. We randomize the response order to mitigate position bias. We use GPT-5-nano \citep{openai2025gpt5} via OpenAI API. We report the \textit{win rate} as the percentage of responses where the judge prefers the model output over the human reference. We also report the \textit{generalization gap} as the difference between source-domain and target-domain win rates. We validate our judge against human annotations on the QA helpfulness task (60 items and 3 annotators per domain). Human--judge agreement ($62$--$80\%$) is comparable to inter-annotator agreement ($69$--$77\%$) across both domains (Appendix~\ref{app:human_validation}). 

\paragraph{Safety evaluation.} For the safety testbed, we evaluate generations with the OpenAI Moderation API \citep{openai2024moderation} and report the \textit{safety score}, defined as $1 - \text{violation\_rate}$, i.e., the fraction of responses not flagged as unsafe. 

\paragraph{Summarization Diversity.}
We measure the linguistic per-input diversity of trained policies for $N=500$ prompts, sampling $K=16$ generations at temperature $T=1.0$, and report the average across all outputs as in \citet{kirk2023understanding}. We assess (i) syntactic diversity via expectation-adjusted distinct (EAD) n-grams ($n=1, \dots, 5$) while applying the length-bias correction proposed by \citet{liu-etal-2022-rethinking}; (ii) semantic diversity via Sentence-BERT \citep[SBERT]{reimers-gurevych-2019-sentence} cosine similarity, defined as one minus the average pairwise cosine similarity between embeddings; and (iii) logical diversity, via natural language inference (NLI) \citep{stasaski-hearst-2022-semantic}, computed as $1 - \mathbb{E}_{(a,b)}\left[p_{\text{ent}}(a,b) - p_{\text{con}}(a,b)\right]$ over sentence pairs $(a,b)$ sampled from the output set, where $p_{\text{ent}}$ and $p_{\text{con}}$ are the entailment and contradiction probabilities assigned by the NLI model. Implementation details are provided in Appendix~\ref{app:hyperparameters}.

\input{latex/table1_with_safety}
\section{Results}

\subsection{Generalization}

Table~\ref{tab:judge_results} presents the win rates, safety scores, and generalization gaps across testbeds.

\paragraph{Pseudo-labeling dominates summarization transfer.}
It reduces cross-model variance by injecting target-domain preference signals. 
For \texttt{Llama-3.1-8B}, pseudo-labeled SFT achieves the highest overall target win rate ($83.37$), while lifting \texttt{OLMo-3-7B} DPO to $72.26$, above all non-synthetic methods. Without pseudo-labeling, adaptation hinges on the SFT stage.

\paragraph{SFT is the key for summarization adaptation.} 
Base models establish a large domain difficulty differential. \texttt{Llama-3.1-8B} scores $44.97$ on TL;DR but only $15.96$ on CNN/DM (gap: $29.01$), while \texttt{OLMo-3-7B} shows a small gap ($2.64$), but a lower baseline performance. SFT reliably reduces the TL;DR$\rightarrow$CNN/DM gap when source and target domain data are included. Source-only SFT improves in-domain performance yet remains brittle. For \texttt{Llama-3.1-8B}, source SFT reaches $36.07$ target win rate, a $+20.11$ gain over the base, but still trails its source win rate by $23.50$. Mix-SFT narrows the gap to $4.25$, a $19.25$ gain over source-only SFT. However, \texttt{OLMo-3-7B} DPO $\mathcal{D}_{S}$ on top of SFT  $\mathcal{D}_{S+T}$  yields $87.78$ on source with a $20.88$ gap, indicating that it can amplify source specialization in a model-dependent way.  

\paragraph{Offline alignment over-specializes to the source domain.}
On summarization, offline methods achieve the highest in-domain win rates but exhibit the largest generalization gaps. On \texttt{OLMo-3-7B}, DPO with target SFT reaches $91.00$ source win rate, yet suffers a $30.60$ gap. On \texttt{Llama-3.1-8B}, source-only DPO reaches $89.87$, yet has a $31.78$ target gap, $10\times$ larger than GRPO ($3.79$). ORPO (gap $25.66$) and direct KTO (gap $38.00$) also show large gaps, consistent with overfitting in-domain.

\paragraph{GRPO prevents domain over-specialization.}
It offers higher source-domain retention than PPO while maintaining comparable cross-domain transfer.
It obtains a $62.57$ source win rate, $+18.27$ over PPO, while keeping the generalization gap at $3.79$. Using target initialization, GRPO remains stable (gap: $-2.99$), avoiding the large gaps of PPO.

\paragraph{PPO underperforms in-domain.}
It produces a large shift toward the target domain on TL;DR$\rightarrow$CNN/DM. 
For \texttt{Llama-3.1-8B}, PPO reaches $59.69$ on target versus $44.30$ on source, surpassing its own source performance. The intermediate SFT model achieves $59.57$ on the source domain, but PPO's online updates cause this to regress to $44.30$, indicating partial forgetting of source cues. Unlike DPO, which memorizes source patterns ($89.87$ on source, $58.09$ on target), its exploration prevents collapse into a ``Reddit'' mode. However, we attribute this partially to source forgetting due to PPO's known training instability rather than to learned domain-general task competence.

\paragraph{QA Helpfulness is largely invariant to domain shift.}
Base models already show near-zero generalization gaps ($-6.78$ for \texttt{Llama-3.1-8B}).
They cluster near zero across all configurations, with DPO with SFT $\mathcal{D}_{S+T}$ initialization yielding a gap of $0.60$.
While summarization generalization gaps span up to $47\%$ across settings, QA win rates remain within a narrow $\pm8$ band, likely because the domain-agnostic helpfulness signals transfer more readily than the stylistic constraints.
However, qualitative inspection (\S\ref{qualitative_analysis}) reveals that models trained on \textit{AskEngineers} often answer culinary questions with engineering-style rigor, which domain-agnostic judges score as helpful despite pragmatic misalignment.
We quantify this effect with a domain-appropriate evaluation in \S\ref{subsec:persona_judge}.

\paragraph{Pseudo-labeling lifts the QA ceiling.}
As in summarization, pseudo-labeling yields the highest absolute QA scores on \texttt{Llama-3.1-8B} ($72$--$76$ target vs ${\sim}60$ for non-synthetic), though the margin is smaller, reflecting the higher baseline.
Online RL methods favor the target domain in QA (GRPO gaps: $-6.66$ to $-7.75$; PPO: $-8.24$), suggesting that exploration-based methods cross near domain shifts more easily than the distant TL;DR$\rightarrow$CNN/DM transfer.

\paragraph{Safety shows moderate domain sensitivity.} 
Base models score below $50\%$, indicating compliance with harmful requests, with model-dependent directionality (\texttt{Llama-3.1-8B} $38.40$/$32.90$, \texttt{OLMo-3-7B} $35.94$/$48.22$). Source SFT provides marginal improvement ($39$--$42$ target), confirming that cybercrime refusal does not transfer to violence prompts. Among online methods, GRPO with a pseudo-labeled RM reaches $99.80$/$95.60$, but with a source RM remain at $43.90$/$39.50$.

\paragraph{KTO peaks at safety without pseudo-labeling.}
Among non-pseudo-labeled methods, direct KTO achieves the highest target safety score on \texttt{OLMo-3-7B} ($76.49\%$; source: $74.52\%$). On the same model, it substantially outperforms direct DPO ($58.12\%$ target; $55.20\%$ source). This may reflect KTO's use of unpaired desirable/undesirable labels derived from PKU-SafeRLHF's pairwise relative-safety annotations.

\paragraph{Pseudo-labeling achieves near-perfect safety.}
On \texttt{Llama-3.1-8B}, all pseudo-labeled objectives (except PPO) achieve over $95\%$ target safety and over $99\%$ on source. The target-domain gap is small, indicating that the teacher's safety prior transfers across harm categories. However, DPO on \texttt{OLMo-3-7B} erases the pseudo-labeled safety floor: SFT on $\mathcal{D}_T^{\text{synth}}$ achieves $99.78$/$95.64$, but adding DPO drops scores to $74.52$/$63.14$. This does not occur with KTO or ORPO on \texttt{OLMo-3-7B} (all $>95\%$), suggesting that DPO's contrastive objective can overwrite SFT safety priors in a model-dependent manner. 

\input{latex/figure2_llama.tex}

\paragraph{Offline and online objectives.} The contrast between offline over-specialization and the comparative robustness of online RL follows from their different training signals. Offline objectives optimize against a fixed set of source preferences, so every step upweights whatever distinguishes them. This produces the largest gaps, as source-only DPO on \texttt{Llama-3.1-8B} reaches $89.87$ in domain but $58.09$ out of it, and ORPO and KTO are similar (gaps $25.66$ and $38.00$). GRPO instead keeps its gap at $3.79$ on the same testbed, an order of magnitude below source-only DPO. Online methods train on rollouts from the current policy, scored by a reward model. The sampling distribution moves with the policy, so no fixed set of source signals is repeated, while the KL anchor bounds deviation from the reference policy. 

Across tasks, the adaptation strategy matters more than the alignment objective: pseudo-labeling is the strongest strategy, and its boost is largest where the domain shift is widest.

\subsection{Diversity}
\label{sub:diversity}

We assess whether generalization gains come at the cost of syntactic, semantic, and logical diversity in summarization across all adaptation and alignment approaches with \texttt{Llama-3.1-8B}  (Figure~\ref{fig:diversity_final_base}).

\input{latex/table_teacher_quality}

\paragraph{Alignment reduces diversity.} 
Shifting from SFT to offline preference objectives contracts syntactic and semantic variety. While source-only SFT maintains the highest semantic diversity ($0.46$), DPO and ORPO drop to $0.24$ and $0.32$. This resilience may stem from training on self-generated rollouts, where exploration and the KL anchor to the reference policy limit convergence to the teacher's templates.

\paragraph{Pseudo-labeling causes mode collapse.} 
Despite high win rates (Table~\ref{tab:judge_results}), pseudo-labeling eliminates semantic and syntactic variety. Semantic diversity drops to near-zero ($0.07$–$0.08$) across offline objectives, and EAD falls from $0.86$ to $0.51$. This suggests a distillation effect in which students overfit the teacher's recurring templates at the expense of the flexibility seen in the SFT adaptations and source-only baselines, consistent with evidence that training on model-generated data can collapse distributional tails \citep{shumailov2024ai}.
From an optimization perspective, pseudo-labeling acts as a form of \textit{distributional regularization} via teacher distillation. By replacing high-variance domain-specific signals with the structured teacher's outputs, the student policy is regularized towards a narrow mode.

\paragraph{Online RL preserves diversity.} Under pseudo-labeling, PPO and GRPO retain higher semantic diversity compared to the offline methods ($0.22$ and $0.25$ vs $0.07$--$0.08$), with GRPO consistently matching or exceeding PPO, slightly outperforming DPO by $0.11$. This resilience likely stems from training on self-generated rollouts, where exploration and the KL anchor to the reference policy limit convergence onto the teacher's templates.

\paragraph{Reduction of logical diversity.} High NLI scores (>$1.0$) indicate contradictions, while lower scores indicate consistency. Pseudo-labeling reduces NLI to $0.88$ (Figure~\ref{fig:diversity_final_base}, Column~3), while SFT-Mix and SFT-Target maintain $1.05 \text{-} 1.10$. For summarization, lower diversity is desirable, suggests consistent factual retrieval, in line with findings in model pruning, that restricted capacity reduces hallucination risk by encouraging higher lexical overlap with the document \citep{chrysostomou-etal-2024-investigating}.  However, this consistency comes at the cost of syntactic and semantic variety (\S\ref{qualitative_analysis}).

\paragraph{Generalization and diversity trade-offs.}
\textit{SFT-Mix} balances syntactic ($0.87$) and semantic ($0.31$) diversity with target generalization. \textit{Pseudo-labeling} maximizes win rates but minimizes diversity, favoring reliability over creative variance. 
\textit{That makes the latter suitable for tasks requiring high reliability but ill-suited for creativity tasks.}

\input{latex/table_persona_aware_judge}

\section{Analysis}
\label{sec:analysis}

\paragraph{Domain Relevance vs. Teacher Quality.}
\label{subsec:teacher_vs_domain}

Target domain pseudo-labeling improves target win rates (Table~\ref{tab:judge_results}). To test whether this stems from the ~\emph{domain relevance} of the synthetic data rather than a generic quality boost from the teacher, we apply the same pseudo-labeling procedure to the \emph{source} domain on the summarization testbed. We produce a synthetic source dataset $\mathcal{D}_S^{\text{synth}}$ and train \texttt{Llama-3.1-8B} under SFT and DPO.
Source-domain pseudo-labeling improves target win rate (Table~\ref{tab:teacher_vs_domain}) by $+8.08$ (SFT) and $+4.79$ (DPO), confirming a generic quality boost from the teacher over the original annotations. However, the generalization gap under source-domain pseudo-labeling ($28.26$ SFT, $27.64$ DPO) is similar to training on the original source data ($23.50$ and $31.78$), i.e., the teacher quality alone does not close the domain gap. Target-domain pseudo-labeling reduces this to $12.33$ (SFT) and $17.29$ (DPO). For DPO, target-domain synthetic data outperforms source-domain synthetic by $+15.62$ points ($78.50$ - $62.88$). The teacher's style is identical in both synthetic conditions, and only the domain of the synthetic data differs. This confirms that domain relevance accounts for the majority of the benefit.

\paragraph{Judge Sensitivity to Epistemic Mismatch.}
\label{subsec:persona_judge}

\input{latex/table2}

A domain-agnostic judge may score responses as helpful even when they reflect the wrong domain's style, e.g., answering culinary questions with engineering-style. We call this an \emph{epistemic mismatch}: the response is logically sound but violates the community's expected expertise and pragmatic conventions. To quantify this, we deploy a domain-aware judge (Appendix~\ref{ref:appendix_prompts}) that penalizes such mismatches. Source-only models lose $3$--$4$ points, while pseudo-labeled DPO \emph{improves} ($+3.48$), confirming genuine domain calibration (Table~\ref{tab:persona_judge}). The near-zero QA generalization gaps in Table~\ref{tab:judge_results} thus partly reflect judge insensitivity. We characterize AskEngineers\,$\rightarrow$\,AskCulinary as a \emph{near domain} shift where surface format is retained but epistemic style and pragmatic persona change.

\paragraph{Data Efficiency of Pseudo-labeling.}
\label{subsec:efficiency}

Strikingly, reducing pseudo-labeled data by $90\%$ in summarization causes only negligible performance drops across all offline objectives (Table~\ref{tab:ablation_dataset_size}). For SFT, KTO, and ORPO, the $10\%$ subset slightly \emph{outperforms} the full dataset target performance, confirming that effectiveness is driven by domain relevance, not volume, as 10\% of the synthetic set suffices to impart target-domain priors. Practically, pseudo-labeling is lightweight, avoiding the cost of scaling teacher generation to full dataset size.

\paragraph{Impact of the SFT Order.}
\label{app:sft_order}

When combining source and target data across multiple SFT stages, the order might play a role. SFT on target ($\mathcal{D}_T$) before the source domain ($\mathcal{D}_S$) achieves a target win rate of $56.40$, compared to $35.22$ when the order is reversed (Table~\ref{tab:curriculum_order}). Establishing target-domain stylistic priors early provides a stable foundation for subsequent task learning. Conversely, late-stage adaptation to the target domain causes drops in summarization competence previously acquired on the source domain. Transitioning from target SFT to source DPO (SFT $\mathcal{D}_T$ $\rightarrow$ DPO $\mathcal{D}_S$) yields a target win rate of $56.82$. Inserting a source SFT step (SFT $\mathcal{D}_T$ $\rightarrow$ SFT $\mathcal{D}_S$ $\rightarrow$ DPO $\mathcal{D}_S$) increases performance to $65.56$, suggesting that this intermediate step realigns the model to the source distribution, before applying DPO.

\input{latex/table3}

\paragraph{Qualitative Analysis.}
\label{qualitative_analysis}

Table~\ref{tab:coloured_output_example} shows an example of epistemic shift in style. Source-only models (\textit{AskEngineers}) adopt an engineering persona that treats cooking as a physical process of heat and fat. In contrast, target adapted models shift toward culinary norms. Both are judged as helpful, as they provide logical justifications, suggesting that judge win rates over-represent structural coherence over the domain-specific vibe or stylistic alignment essential for true expert-level transfer. Full examples are in Appendix~\ref{app:example}.

\paragraph{Structural mode collapse.}
While pseudo-labeling achieves the highest target win rates, it introduces a characteristic failure mode. The pseudo-labeled model maps topically unrelated \textit{CNN/DM} articles onto a single template, opening with ``\textit{The article discusses}'' and proceeding through ``\textit{The author argues/advises/describes}'' to ``\textit{The article also recommends}'' (Table~\ref{tab:collapse_main}). The source-only (\textit{TL;DR}) DPO model, by contrast, preserves distinct editorial voices. For example, a direct and opinionated advisory tone for the food trends piece, a warm personal anecdote for the travel article, and a factual, comparative framing for the sports commentary. Quantifying the patterns by exact matching, this frame of discourse markers appears in ${\sim}10\%$ of pseudo-labeled summaries against only ${\sim}0.2\%$ of source trained outputs and human reference summaries, closely mirroring the teacher's own rate ($10.5\%$). The student inherits the teacher's discourse style rather than the target domain's conventions. Because each summary remains fluent and structurally coherent, the collapse is invisible to the LLM judge, and win rates are maximized where tonal variation is lost. Full examples are in Appendix~\ref{app:pseudo_failures}.

\input{latex/table_qual_main_body}

\paragraph{Deployment Guidance.}
The optimal deployment strategy depends on the reliability and diversity trade-off. 
For \textbf{high-constraint tasks} (e.g., news summarization), pseudo-labeling on target paired with offline methods maximizes target win rates while producing factually self-consistent outputs.
For \textbf{creative or open-ended tasks}, where output variety matters, $\mathcal{D}_{S+T}$ SFT preserves diversity while maintaining competitive generalization. Among preference objectives, GRPO offers the best diversity and generalization balance. For \textbf{safety critical tasks}, pseudo-labeling achieves near-perfect refusal rates across harm categories; among non-synthetic methods, KTO offers the best safety and robustness balance.

\input{latex/table_collapse_main_body}

\section{Conclusion \& Takeaways} 
\label{sec:conclusion}

We presented a systematic study of preference-optimization under domain shift. First, the adaptation strategy is more influential than the alignment objective, exposing a critical vulnerability, offline alignment methods are highly brittle under domain shift and fail to transfer without explicit target-domain adaptation. Second, synthetic supervision is a double-edged sword. While pseudo-labeling yields the highest target win rates, it induces severe mode collapse. This diversity tax results in models that are highly reliable but linguistically monotonous. 
Future work should explore hybrid data mixtures and move beyond scalar win rates to optimize for diversity, maximizing generalization without collapsing into single-mode distributions. Additionally, we will investigate how instruction and label noise impacts alignment generalization under domain shift \citep{alajrami-etal-2025-fine} and extend our analysis to cross-lingual settings using unlabeled target data \citep{yamaguchi2025adapting}.

\section*{Limitations}
\label{sec:limitations}

First, while the models we experiment with are representative of standard deployment, larger frontier models may exhibit different generalization dynamics or resistance to forgetting. Second, we focus solely on the English language. Reasoning-intensive tasks (e.g., coding) or multilingual settings rely on different internal mechanisms and may manifest the ``alignment tax'' differently. Third, each testbed evaluates a single transfer direction. Finally, the pseudo-labeling strategy relies on a stronger ``teacher'' model. Synthetic preferences cannot guarantee perfect alignment with human intent; teacher hallucinations or biases are inevitably distilled into the student, potentially causing the observed mode collapse.

\section*{Ethical Considerations}
\label{sec:ethics}

The trade-off between alignment performance and output diversity carries significant ethical implications. We demonstrate that while pseudo-labeling improves domain transfer, it induces severe mode collapse. Deployment of such models risks homogenizing machine-generated content, reducing ``cognitive diversity'' in creative or exploratory applications. Furthermore, we caution against uncritical reliance on models adapted via synthetic loops. The student model may amplify the latent biases of the teacher. In high-stakes domains, this risks generating confident outputs that mimic the target domain's style but lack factual grounding. Benchmark performance alone is insufficient justification for deployment without rigorous human-in-the-loop verification.  Finally, the safety testbed uses data from PKU-SafeRLHF \citep{ji-etal-2025-pku}, a public dataset designed for safety research. Model outputs were evaluated via automated moderation but were not deployed or made available.

\section*{Acknowledgments}

We thank Jasivan Sivakumar, Vynska Amalia Permadi, Sam Lewis-Lim, Yanwen Peng, and Atsuki Yamaguchi for their valuable help and feedback. CK is supported by the Centre for Doctoral Training in Speech and Language Technologies (SLT) and their Applications funded by UK Research and Innovation grant [grant number EP/S023062/1]. XT and NA are supported by the EPSRC [grant number EP/Y009800/1], through funding from Responsible AI UK (KP0016) as a Keystone project. We acknowledge (1) IT Services at the University of Sheffield for the provision of services for high-performance computing; (2) the use of the University of Oxford Advanced Research Computing (ARC) facility; (3) the EuroHPC Joint Undertaking for awarding this project access to the EuroHPC supercomputer LEONARDO, hosted by CINECA (Italy) and the LEONARDO consortium through a EuroHPC Development Access call; (4) the use of resources provided by the Isambard-AI National AI Research Resource (AIRR). Isambard-AI is operated by the University of Bristol and is funded by the UK Government’s Department for Science, Innovation and Technology (DSIT) via UK Research and Innovation; and the Science and Technology Facilities Council [ST/AIRR/I-A-I/1023].

\sloppy
\bibliography{custom,anthology-1,anthology-2}

\clearpage

\appendix

\section{Implementation Details}
\label{app:implementation}

\subsection{Training and Optimization Hyperparameters}
\label{app:hyperparameters}

All experiments use Low-Rank Adaptation \citep[LoRA]{hu2022lora} and a shared optimization setup. Table~\ref{tab:hyperparams} lists the hyperparameters.
We generate synthetic preferences using \texttt{Llama-3.3-70B-Instruct} \citep{grattafiori2024llama3herdmodels}. 
We use \texttt{gpt-5-nano-2025-08-07} for LLM-as-a-judge evaluation and \texttt{omni-moderation-latest} for safety evaluation. For diversity analysis, semantic diversity uses \texttt{all-mpnet-base-v2} \citep{reimers-gurevych-2019-sentence}, while logical diversity uses \texttt{roberta-large-mnli} \citep{liu2019robertarobustlyoptimizedbert}.

\input{latex/table_hyperparams}

\subsection{Decoding and Generation}

All evaluations use temperature sampling with temperature $0.7$ and top-$p=0.9$. Maximum generation length is dataset-dependent: 128 tokens for helpfulness and safety, and up to 1024 tokens for summarization.  
For diversity analysis, we sample $K=16$ generations per prompt at temperature $1.0$.

\subsection{Reproducibility}

We fix random seeds at the framework, data-loader, and model levels. Results correspond to the final training checkpoint.

\subsection{Software}

We use \texttt{transformers} \citep{wolf-etal-2020-transformers}, \texttt{trl} \citep{vonwerra2020trl}, for SFT, DPO, KTO, ORPO, PPO, GRPO, and reward modeling training, \texttt{peft} \citep{peft} for LoRA adaptation, and \texttt{vllm} for generation \citep{kwon2023efficient}.

\section{Licences}
\label{app:licences}

We use publicly available models and datasets under the licences listed below.

\subsection{Models}
\begin{sloppypar}
\texttt{Llama-3.3-70B-Instruct} and \texttt{Llama-3.1-8B} are distributed under the Llama 3.3 and Llama 3.1 Community Licence Agreements, respectively.\footnote{\url{https://www.llama.com/llama3_1/license/}} \texttt{OLMo-3-7B} is distributed under the Apache Licence 2.0.\footnote{\url{https://github.com/allenai/OLMo-core/blob/main/LICENSE}} \texttt{gpt-5-nano} (LLM-as-a-judge) and OpenAI Moderation API  \texttt{omni-moderation-latest} (safety evaluation) are accessed via the OpenAI API and governed by the OpenAI Terms of Use \footnote{\url{https://openai.com/policies/row-terms-of-use/}}; the Moderation endpoint is provided free of charge.\footnote{\url{https://platform.openai.com/docs/guides/moderation}}  For diversity evaluation, we use \texttt{all-mpnet-base-v2} (Apache 2.0) and \texttt{roberta-large-mnli} (MIT).
\end{sloppypar}

\subsection{Datasets}

CNN/DailyMail is distributed under the Apache Licence 2.0.\footnote{\url{https://huggingface.co/datasets/abisee/cnn_dailymail}} The Reddit TL;DR corpus \citep{volske-etal-2017-tl} is distributed under CC BY 4.0;\footnote{\url{https://zenodo.org/records/1043504}} the associated human feedback annotations \citep{stiennon2020learning} are released for research use under the MIT licence.\footnote{\url{https://github.com/openai/summarize-from-feedback}} The Stanford Human Preferences dataset \citep{pmlr-v162-ethayarajh22a} is distributed under the MIT licence.\footnote{\url{https://huggingface.co/datasets/stanfordnlp/SHP}} PKU-SafeRLHF \citep{ji-etal-2025-pku} is distributed under CC BY-NC 4.0.\footnote{\url{https://huggingface.co/datasets/PKU-Alignment/PKU-SafeRLHF}}

\section{Dataset Details}
\input{latex/table_dataset_stats}

We use the source domains ($\mathcal{D}_S$) to provide human-labeled preference signals for initial alignment. We use the target domains ($\mathcal{D}_T$) to facilitate adaptation via target SFT and pseudo-labeling (\S\ref{subsec:adaptation_strategies}). Evaluation occurs on held-out splits of both domains to measure the generalization gap.

\section{Pseudo-Labeling Prompts}
\label{app:pseudo_prompts}

To generate the pseudo-labeled dataset $\mathcal{D}_T^{\text{synth}}$, we use \texttt{Llama-3.3-70B-Instruct} as the teacher model. For each target-domain prompt, we generate $k=3$ candidates using temperature $0.7$ and top-$p=0.9$. The maximum generation length is 512 tokens. Every nonempty teacher candidate is retained and paired as the preferred response $y_w$, with the original response as the dispreferred response $y_l$. Table~\ref{tab:pseudo_label_prompts} provides the task prompts.

\paragraph{Validation of synthetic preference pairs.}
To validate our assumption that teacher-generated responses provide a superior structural signal compared to noisy human references, we evaluate 200 synthetic target-domain pairs ($\mathcal{D}_T^{\text{synth}}$) using our \texttt{gpt-5-nano} judge. The teacher-generated candidate ($y_w$) was preferred over the original ground-truth reference ($y_l$) in $96.50\%$ of the cases for CNN/DM and in $97.50\%$ of the cases for AskCulinary, confirming that pairing these responses creates a robust hard negative signal that effectively denoises the preference distribution.

\input{latex/table_pseudo_prompts}

\section{LLM-as-a-Judge Prompts}
\label{ref:appendix_prompts}

We use \texttt{gpt-5-nano} as an LLM judge. The model receives a prompt, a reference response (e.g., the ground-truth summary or the \texttt{chosen} response from the test set), and a candidate response generated by one of our fine-tuned models. The judge's task is to determine which response is better. The order of the reference and candidate responses is randomized to mitigate position bias. Table \ref{tab:judge_prompts} contains the specific prompts used. The Helpfulness Domain Aware prompt corresponds to the judge sensitivity analysis of \S\ref{subsec:persona_judge}.

\section{Human Validation of the LLM-as-a-Judge}
\label{app:human_validation}

To validate the reliability of our LLM-based evaluation, we conduct a human annotation study on the QA helpfulness task. LLM-as-a-judge evaluation has become standard practice in alignment research \citep{zheng2023judging, liu-etal-2023-g}, and achieves agreement with human preferences comparable to inter-human agreement \citep{zheng2023judging}. Nevertheless, task-specific validation remains important, as judge reliability can vary across domains and evaluation criteria \citep{thakur-etal-2025-judging}. \citet{rafailov2023direct} and \citet{kirk2023understanding} run similar human validation studies for the summarization setting against a GPT-4 LLM-as-a-Judge.

\paragraph{Setup.}
We sample 60 prediction--reference pairs per domain, from AskEngineers and AskCulinary, stratified across experimental conditions spanning the adaptation spectrum.
Each pair is presented to human annotators in randomized A/B order, with the question: \emph{``Given a user post and two responses, which response is more helpful to the user?''}, the same criterion used by the automated judge.
Annotators select from three options: Response~A is better, Response~B is better, or Tie.
We recruit five volunteer annotators, assigning three per domain. Annotators receive written guidelines and a browser-based annotation interface. No domain expertise was required.

\input{latex/table_human_val}

\paragraph{Results.}
Table~\ref{tab:human_judge_agreement} reports human--judge agreement, inter-annotator agreement (IAA), and Cohen's $\kappa$ \citep{cohen1960coefficient}.  Both agreement and $\kappa$ are computed on non-tie items. On AskEngineers, human--judge agreement ranges from 62--78\%, with average inter-annotator agreement of 69.3\%. On AskCulinary, human--judge agreement ranges from 62--80\%, with average inter-annotator agreement of 77.2\%. In both domains, human--judge agreement falls within the range of inter-annotator agreement, consistent with the finding that LLM judges achieve parity with human preferences \citep{zheng2023judging, rafailov2023direct}.

\section{Diversity Analysis}

Table~\ref{tab:diversity_metrics} provides the numerical results for the diversity analysis discussed in \S\ref{sub:diversity}.
\input{latex/table_diversity}

\input{latex/table_prompts}

\section{Qualitative Case Studies}
\label{app:both_examples}
\subsection{Epistemic Drift}
\label{app:example}
Table~\ref{tab:qualitative_dpo_appendix_colour} presents a case study of persona shift in QA helpfulness during the transfer from \textit{AskEngineers} to \textit{AskCulinary}. Models aligned only on source data (\textit{AskEngineers}) yield an engineering persona that treats cooking as a physical process of heat and fat management. In contrast, target-adapted models successfully shift toward culinary norms, adopting a culinary persona.  These examples illustrate the epistemic drift where models adapted to the target domain adopt culinary reasoning, whereas source-only models maintain an engineering persona even when providing helpful cooking advice. While different, both are judged as helpful because they provide logical justifications. This suggests that LLM-as-a-judge win rates may over-represent structural coherence while under-representing the domain specific vibe or stylistic alignment essential for true expert-level transfer.

\input{latex/table_qual_colour}

\subsection{Pseudo-Labeling Mode Collapse}
\label{app:pseudo_failures}

Table~\ref{tab:pseudo_failures} illustrates the mode collapse on the summarization target domain.
We compare DPO trained on source preferences ($\mathcal{D}_S$) against DPO trained on pseudo-labeled target data ($\mathcal{D}_T^{\text{synth}}$), both using \texttt{Llama-3.1-8B}.
Despite achieving the highest win rates (Table~\ref{tab:judge_results}), the pseudo-labeled model exhibits \textit{structural mode collapse}: it maps articles on entirely unrelated topics to the same template. All three examples, covering food trends, Greek island tourism, and football rules, are reduced to an identical skeleton: ``\textit{The article discusses~\ldots; The author argues/advises/describes~\ldots; The article also recommends~\ldots}''.

\input{latex/table_pseudo_failures}

\end{document}

%% file: latex/figure_final.tex
\begin{figure}[t]
\centering
\includegraphics[width=\columnwidth]{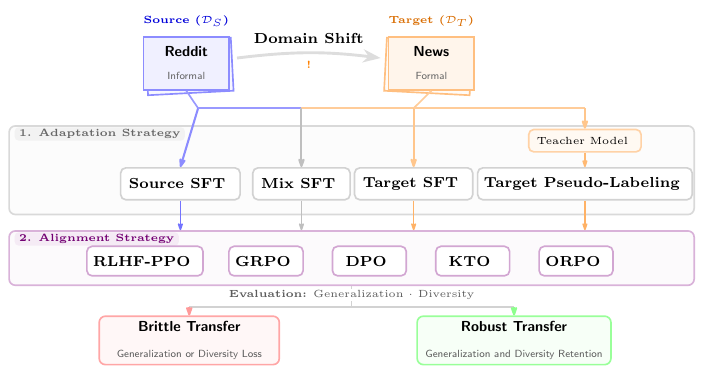}
\caption{We decompose preference optimization under domain shift into: \textbf{Adaptation} and \textbf{Alignment}.}
\label{fig:framework}
\end{figure}

%% file: latex/table1_with_safety.tex
\begin{table*}[t]
\centering
\setlength{\tabcolsep}{2pt}
\resizebox{\textwidth}{!}{%
\tiny
\begin{tabular}{l cc ccc ccc ccc ccc ccc ccc}
\toprule
 & & &
 \multicolumn{6}{c}{\textbf{Summarization} ($\uparrow$)} &
 \multicolumn{6}{c}{\textbf{QA Helpfulness} ($\uparrow$)} &
 \multicolumn{6}{c}{\textbf{Safety} ($\uparrow$)} \\
\cmidrule(lr){4-9} \cmidrule(lr){10-15} \cmidrule(lr){16-21}

 & \multicolumn{2}{c}{\textbf{Data}} &
 \multicolumn{3}{c}{\textit{Llama-3.1-8B}} &
 \multicolumn{3}{c}{\textit{OLMo-3-7B}} &
 \multicolumn{3}{c}{\textit{Llama-3.1-8B}} &
 \multicolumn{3}{c}{\textit{OLMo-3-7B}} &
 \multicolumn{3}{c}{\textit{Llama-3.1-8B}} &
 \multicolumn{3}{c}{\textit{OLMo-3-7B}} \\
\cmidrule(lr){2-3} \cmidrule(lr){4-6} \cmidrule(lr){7-9}
\cmidrule(lr){10-12} \cmidrule(lr){13-15}
\cmidrule(lr){16-18} \cmidrule(lr){19-21}

\textbf{Method} & \textbf{SFT} & \textbf{Pref.} &
 Src & Tgt & Gap &
 Src & Tgt & Gap &
 Src & Tgt & Gap &
 Src & Tgt & Gap &
 Src & Tgt & Gap &
 Src & Tgt & Gap \\
\midrule

\textbf{Base} & -- & --
 & 44.97 & 15.96 & \posL 29.01
 & 41.78 & 39.14 & \posS 2.64
 & 54.59 & 61.37 & \negS -6.78
 & 60.67 & 57.34 & \posS 3.33
 & 38.40 & 32.90 & \posS 5.50
 & 35.94 & 48.22 & \negL -12.28 \\
\midrule

\rowcolor{gray!30}
\multicolumn{1}{l}{\cellcolor{white}\multirow{4}{*}{\rotatebox{90}{\textbf{SFT}}}} &  $\mathcal{D}_S$ & --
 & 59.57 & 36.07 & \posL 23.50
 & 43.09 & 39.04 & \posS 4.05
 & 60.74 & 60.08 & \posS 0.66
 & 61.30 & 64.35 & \negS -3.05
 & 43.40 & 39.20 & \posS 4.20
 & 42.31 & 42.40 & \negS -0.09 \\

 & $\mathcal{D}_{S+T}$ & --
 & 61.56 & 57.31 & \posS 4.25
 & 41.50 & 40.40 & \posS 1.10
 & 59.14 & 60.68 & \negS -1.54
 & 62.33 & 66.16 & \negS -3.83
 & 42.80 & 40.20 & \posS 2.60
 & 42.53 & 38.71 & \posS 3.82 \\

 &  $\mathcal{D}_T$ & --
 & 66.20 & 54.90 & \posL 11.30
 & 39.58 & 38.24 & \posS 1.34
 & 63.81 & 64.94 & \negS -1.13
 & 61.72 & 66.68 & \negS -4.96
 & 42.20 & 40.70 & \posS 1.50
 & 44.07 & 40.82 & \posS 3.25 \\

 & $\mathcal{D}_T^{\text{synth}}$ & --
 & \underline{95.70} & \underline{\textbf{83.37}} & \posL 12.33
 & \underline{75.16} & \underline{70.54} & \posS 4.62
 & \underline{\textbf{72.79}} & \underline{76.04} & \negS -3.25
 & \underline{72.23} & \underline{66.74} & \posS 5.49
 & \underline{99.60} & \underline{95.90} & \posS 3.70
 & \underline{\textbf{99.78}} & \underline{95.64} & \posS 4.14 \\
\midrule

\rowcolor{gray!30}
\multicolumn{1}{l}{\cellcolor{white}}
 &  $\mathcal{D}_S$ &  $\mathcal{D}_S$
 & 89.87 & 58.09 & \posL 31.78
 & 40.74 & 41.70 & \negS -0.96
 & 60.73 & 61.45 & \negS -0.72
 & 60.25 & 57.53 & \posS 2.72
 & 44.00 & 42.00 & \posS 2.00
 & 78.99 & \underline{67.37} & \posL 11.62 \\
\rowcolor{gray!30}
\multicolumn{1}{l}{\cellcolor{white}} & -- &  $\mathcal{D}_S$
 & 85.17 & 38.29 & \posL 46.88
 & 41.10 & 40.10 & \posS 1.00
 & 64.01 & 61.96 & \posS 2.05
 & 60.02 & 56.69 & \posS 3.33
 & 41.90 & 41.90 & 0.00
 & 55.20 & 58.12 & \negS -2.92 \\

 & $\mathcal{D}_{S+T}$ &  $\mathcal{D}_S$
 & 87.72 & 68.50 & \posL 19.22
 & 87.78 & 66.90 & \posL 20.88
 & 59.40 & 58.80 & \posS 0.60
 & 59.07 & 57.40 & \posS 1.67
 & 42.20 & 43.10 & \negS -0.90
 & \underline{82.72} & 65.65 & \posL 17.07 \\

 & $\mathcal{D}_T$ &  $\mathcal{D}_S$
 & 67.83 & 56.82 & \posL 11.01
 & \underline{\textbf{91.00}} & 60.40 & \posL 30.60
 & 59.29 & 64.69 & \negS -5.40
 & 60.21 & 58.15 & \posS 2.06
 & 44.40 & 43.50 & \posS 0.90
 & 74.89 & 60.77 & \posL 14.12 \\
\multirow{-5}{*}{\rotatebox{90}{\textbf{DPO}}}
 & $\mathcal{D}_T^{\text{synth}}$ & $\mathcal{D}_T^{\text{synth}}$
 & \underline{95.79} & \underline{78.50} & \posL 17.29
 & 80.16 & \underline{\textbf{72.26}} & \posS 7.90
 & \underline{72.76} & \underline{75.52} & \negS -2.76
 & \underline{63.59} & \underline{65.27} & \negS -1.68
 & \underline{\textbf{99.80}} & \underline{\textbf{96.40}} & \posS 3.40
 & 74.52 & 63.14 & \posL 11.38 \\
\midrule

\rowcolor{gray!30}
\multicolumn{1}{l}{\cellcolor{white}}
 & -- &  $\mathcal{D}_S$
 & 79.00 & 41.00 & \posL 38.00
 & 41.40 & 40.00 & \posS 1.40
 & 64.22 & 61.53 & \posS 2.69
 & 61.25 & 56.99 & \posS 4.26
 & 71.50 & 70.50 & \posS 1.00
 & 74.52 & 76.49 & \negS -1.97 \\
\rowcolor{gray!30}
\multicolumn{1}{l}{\cellcolor{white}}
 &  $\mathcal{D}_S$ &  $\mathcal{D}_S$
 & 81.06 & 51.10 & \posL 29.96
 & 40.88 & 39.64 & \posS 1.24
 & 61.03 & 58.95 & \posS 2.08
 & 62.70 & 58.55 & \posS 4.15
 & 33.30 & 35.50 & \negS -2.20
 & 65.74 & 61.56 & \posS 4.18 \\

 & $\mathcal{D}_{S+T}$ &  $\mathcal{D}_S$
 & 60.92 & 55.40 & \posS 5.52
 & 77.35 & 62.06 & \posL 15.29
 & 62.17 & 66.29 & \negS -4.12
 & 55.39 & 54.70 & \posS 0.69
 & 67.60 & 64.10 & \posS 3.50
 & 70.94 & 64.99 & \posS 5.95 \\

 & $\mathcal{D}_T$ &  $\mathcal{D}_S$
 & 65.05 & 56.41 & \posS 8.64
 & 77.38 & 58.70 & \posL 18.68
 & 63.23 & 58.29 & \posS 4.94
 & 56.26 & 54.66 & \posS 1.60
 & 66.00 & 60.90 & \posS 5.10
 & 71.45 & 59.31 & \posL 12.14 \\
\multirow{-5}{*}{\rotatebox{90}{\textbf{KTO}}}
 & $\mathcal{D}_T^{\text{synth}}$ & $\mathcal{D}_T^{\text{synth}}$
 & \underline{95.37} & \underline{83.01} & \posL 12.36
 & \underline{78.30} & \underline{70.16} & \posS 8.14
 & \underline{72.39} & \underline{75.38} & \negS -2.99
 & \underline{63.10} & \underline{66.04} & \negS -2.94
 & \underline{\textbf{99.80}} & \underline{95.50} & \posS 4.30
 & \underline{\textbf{99.78}} & \underline{95.77} & \posS 4.01 \\
\midrule

\rowcolor{gray!30}
\rowcolor{gray!30}
\multicolumn{1}{l}{\cellcolor{white}} & -- &  $\mathcal{D}_S$
 & 64.22 & 47.60 & \posL 16.62
 & 40.10 & 40.74 & \negS -0.64
 & 53.66 & 51.33 & \posS 2.33
 & 59.93 & 57.46 & \posS 2.47
 & 53.50 & 46.90 & \posS 6.60
 & 47.00 & 42.67 & \posS 4.33 \\
\rowcolor{gray!30}
\multicolumn{1}{l}{\cellcolor{white}}
 &  $\mathcal{D}_S$ &  $\mathcal{D}_S$
 & 60.96 & 35.30 & \posL 25.66
 & 67.27 & 54.00 & \posL 13.27
 & 54.14 & 58.34 & \negS -4.20
 & 53.34 & 48.08 & \posS 5.26
 & 54.80 & 47.20 & \posS 7.60
 & 47.66 & 41.88 & \posS 5.78 \\

 & $\mathcal{D}_{S+T}$ &  $\mathcal{D}_S$
 & 60.76 & 57.03 & \posS 3.73
 & 65.07 & 56.60 & \posS 8.47
 & 53.82 & 59.72 & \negS -5.90
 & 50.11 & 53.73 & \negS -3.62
 & 53.10 & 46.60 & \posS 6.50
 & 46.12 & 39.63 & \posS 6.49 \\

 & $\mathcal{D}_T$ &  $\mathcal{D}_S$
 & 64.99 & 57.10 & \posS 7.89
 & 59.08 & 43.74 & \posL 15.34
 & 58.83 & 54.92 & \posS 3.91
 & 57.67 & 60.08 & \negS -2.41
 & 52.00 & 47.00 & \posS 5.00
 & 46.12 & 39.37 & \posS 6.75 \\
\multirow{-5}{*}{\rotatebox{90}{\textbf{ORPO}}}
 & $\mathcal{D}_T^{\text{synth}}$ & $\mathcal{D}_T^{\text{synth}}$
 & \underline{\textbf{96.80}} & \underline{82.38} & \posL 14.42
 & \underline{76.17} & \underline{71.45} & \posS 4.72
 & \underline{72.75} & \underline{\textbf{76.15}} & \negS -3.40
 & \underline{\textbf{72.90}} & \underline{65.82} & \posS 7.08
 & \underline{99.60} & \underline{95.50} & \posS 4.10
 & \underline{99.41} & \underline{\textbf{96.30}} & \posS 3.11 \\
\midrule

\rowcolor{gray!30}
\multicolumn{1}{l}{\cellcolor{white}\multirow{5}{*}{\rotatebox{90}{\textbf{PPO}}}} &  $\mathcal{D}_S$ &  $\mathcal{D}_S$
 & 44.30 & 59.69 & \negL -15.39
 & \underline{48.21} & 41.70 & \posS 6.51
 & 55.10 & 58.05 & \negS -2.95
 & 60.98 & 58.15 & \posS 2.83
 & \underline{40.19} & \underline{46.24} & \negS -6.05
 & 35.94 & 48.22 & \negL -12.28  \\

 & $\mathcal{D}_{S+T}$ &  $\mathcal{D}_S$
 & 62.50 & 58.10 & \posS 4.40
 & 46.28 & 42.92 & \posS 3.36
 & 55.50 & 61.39 & \negS -5.89
 & 62.90 & 61.72 & \posS 1.18
 & 39.82 & 43.99 & \negS -4.17
 & 36.60 & \underline{49.54} & \negL -12.94 \\

 & $\mathcal{D}_T$ &  $\mathcal{D}_S$
 & 45.10 & 60.14 & \negL -15.04
 & 46.41 & 43.00 &  \posS 3.41
 & 56.81 & 65.05 &  \negS -8.24
 & 57.56 & 65.53 & \negS -7.97
 & 39.68 & 43.99 & \negS -4.31
 & \underline{36.75} & 48.61 & \negL -11.86 \\

 & $\mathcal{D}_T^{\text{synth}}$ & $\mathcal{D}_T^{\text{synth}}$
 & \underline{71.87} & \underline{61.42} & \posL 10.45
 & 47.52 & \underline{60.92} & \negL -13.40
 & \underline{67.80} & \underline{72.45} & \negS -4.65
 & \underline{72.84} & \underline{\textbf{68.50}} & \posS 4.34
 & 38.80 & 44.52 & \negS -5.72
 & 36.53 & 48.88 & \negL -12.35 \\
 
\midrule

\rowcolor{gray!30}
\multicolumn{1}{l}{\cellcolor{white}\multirow{5}{*}{\rotatebox{90}{\textbf{GRPO}}}} &  $\mathcal{D}_S$ &  $\mathcal{D}_S$
 & 62.57 & 58.78 & \posS 3.79
 & 51.10 & 42.80 & \posS 8.30
 & 54.89 & 62.00 & \negS -7.11
 & 61.45 & 58.30 & \posS 3.15
 & 43.90 & 39.50 & \posS 4.40
 & 46.34 & 46.63 & \negS -0.29 \\

 & $\mathcal{D}_{S+T}$ &  $\mathcal{D}_S$
 & 67.94 & 60.74 & \posS 7.20
 & 72.45 & 55.87 & \posL 16.58
 & 54.60 & 61.26 & \negS -6.66
 & 62.58 & 60.15 & \posS 2.43
 & 41.90 & 40.40 & \posS 1.50
 & 44.95 & 42.54 & \posS 2.41 \\

 & $\mathcal{D}_T$ &  $\mathcal{D}_S$
 & 60.10 & 63.09 & \negS -2.99
 & 68.05 & 52.14 & \posL 15.91
 & 54.40 & 62.15 & \negS -7.75
 & 60.24 & 63.50 & \negS -3.26
 & 43.60 & 41.50 & \posS 2.10
 & 50.44 & 43.59 & \posS 6.85 \\

 & $\mathcal{D}_T^{\text{synth}}$ & $\mathcal{D}_T^{\text{synth}}$
 & \underline{87.16} & \underline{80.19} & \posS 6.97
 & \underline{73.45} & \underline{68.89} & \posS 4.56
 & \underline{64.63} & \underline{62.39} & \posS 2.24
 & \underline{64.50} & \underline{65.10} & \negS -0.60
 & \underline{\textbf{99.80}} & \underline{95.60} & \posS 4.20
 & \underline{\textbf{99.78}} & \underline{96.04} & \posS 3.74 \\

\bottomrule
\end{tabular}
}
\caption{Source and target domain performance across three testbeds.
Summarization and QA report LLM-as-a-judge win rates (\%); Safety reports
safety score (\%). \textbf{Bold}: best score per domain-model pair. Gap $=$ Src $-$ Tgt.}
\label{tab:judge_results}
\end{table*}

%% file: latex/figure2_llama.tex
\definecolor{colorGray}{HTML}{CCCCCC}   
\definecolor{colorCyan}{HTML}{76D7EA}   
\definecolor{colorBlue}{HTML}{4E84AA}   
\definecolor{colorPurple}{HTML}{B24080} 
\definecolor{colorPink}{HTML}{F78181}   
\definecolor{colorGreen}{HTML}{66CC99}  

\pgfplotsset{
    nli style/.style={
        ymin=0.7, ymax=1.20,
        ytick={0.7, 0.8, 0.9, 1.0, 1.1},
    }
}

\begin{figure}[t!] 
    \centering
    \resizebox{\columnwidth}{!}{%
    \begin{tikzpicture}
        \newcommand{\addSixBars}[6]{
            \addplot[fill=colorGray]   coordinates {(0,#1)};
            \addplot[fill=colorCyan]   coordinates {(0,#2)};
            \addplot[fill=colorBlue]   coordinates {(0,#3)};
            \addplot[fill=colorPurple] coordinates {(0,#4)};
            \addplot[fill=colorPink]   coordinates {(0,#5)};
            \addplot[fill=colorGreen]  coordinates {(0,#6)};
        }

        \begin{groupplot}[
            group style={
                group size=3 by 4, 
                horizontal sep=0.4cm,
                vertical sep=0.4cm,
                y descriptions at=edge left,
            },
            area legend,
            ybar=2.4pt,
            width=0.32\textwidth,
            height=3.8cm,
            ymin=0,
            axis on top,
            tick align=inside,
            axis line style={black},
            xtick align=inside,
            ytick align=inside,
            ymajorgrids=true,
            grid style={dashed, black!30},
            xmin=-0.5, xmax=0.5,
            xtick=\empty,
            xticklabels={}, 
            yticklabel style={
                font=\tiny, 
                /pgf/number format/fixed,
                /pgf/number format/precision=1,
                /pgf/number format/fixed zerofill
            },
            ylabel style={font=\normalsize\bfseries, align=center, yshift=-0.1cm},
            every axis plot/.append style={
                bar width=13pt, 
                draw=none
            },
            legend style={
                legend columns=-1,
                font=\footnotesize\sffamily,
                draw=black,
                fill=white,
                at={(0.5,-0.2)},
                anchor=north,
                /tikz/every even column/.append style={column sep=0.3cm},
            },
            legend image code/.code={%
             \draw[#1] (0cm,-0.1cm) rectangle (0.2cm,0.2cm);
            }
        ]

        \nextgroupplot[title={\textbf{Syntactic (EAD)}}, ylabel=\textbf{Source-Only}, ymax=1.0, ytick={0,0.25,0.5,0.75,1}, legend to name=GlobalLegend]
        \addplot[fill=colorGray]  coordinates {(0,0.8644)}; \addlegendentry{SFT}
        \addplot[fill=colorCyan]  coordinates {(0,0.63)};   \addlegendentry{PPO}
        \addplot[fill=colorBlue]  coordinates {(0,0.68)};   \addlegendentry{GRPO}
        \addplot[fill=colorPurple] coordinates {(0,0.75)};   \addlegendentry{DPO}
        \addplot[fill=colorPink]   coordinates {(0,0.7618)}; \addlegendentry{KTO}
        \addplot[fill=colorGreen]  coordinates {(0,0.7567)}; \addlegendentry{ORPO}

        \nextgroupplot[title={\textbf{Semantic (SBERT)}}, ymax=0.5, ytick={0,0.1,0.2,0.3,0.4,0.5}]
        \addSixBars{0.4643}{0.21}{0.35}{0.2353}{0.4073}{0.3248}

        \nextgroupplot[title={\textbf{Logical (NLI)}}, nli style]
        \addSixBars{1.0773}{1.07}{1.0756}{0.9485}{1.0677}{1.0067}

        \nextgroupplot[ylabel={Target-SFT}, ymax=1.0, ytick={0,0.25,0.5,0.75,1}]
        \addSixBars{0.7702}{0.77}{0.75}{0.7700}{0.7705}{0.7712}

        \nextgroupplot[ymax=0.5, ytick={0,0.1,0.2,0.3,0.4,0.5}]
        \addSixBars{0.2979}{0.31}{0.31}{0.2968}{0.2972}{0.2981}

        \nextgroupplot[nli style]
        \addSixBars{1.1004}{1.0962}{1.0993}{1.0996}{1.0995}{1.0973}

        \nextgroupplot[ylabel={Mix-SFT}, ymax=1.0, ytick={0,0.25,0.5,0.75,1}]
        \addSixBars{0.8726}{0.76}{0.76}{0.7232}{0.7716}{0.7718}

        \nextgroupplot[ymax=0.5, ytick={0,0.1,0.2,0.3,0.4,0.5}]
        \addSixBars{0.3062}{0.2465}{0.2985}{0.2074}{0.3064}{0.3063}

        \nextgroupplot[nli style]
        \addSixBars{1.0974}{1.0743}{1.093}{1.0474}{1.0973}{1.0950}

        \nextgroupplot[ylabel={Pseudo-Label}, ymax=1.0, ytick={0,0.25,0.5,0.75,1}]
        \addSixBars{0.5122}{0.62}{0.62}{0.5163}{0.5118}{0.5258}

        \nextgroupplot[ymax=0.5, ytick={0,0.1,0.2,0.3,0.4,0.5}]
        \addSixBars{0.0726}{0.2204}{0.2484}{0.0657}{0.0730}{0.0761}

        \nextgroupplot[nli style]
        \addSixBars{0.8835}{0.9256}{0.9456}{0.8918}{0.8859}{0.8939}

        \end{groupplot}
        
    \end{tikzpicture}%
    }
    \vspace{0.1cm}
    \centerline{\resizebox{0.7\linewidth}{!}{\ref*{GlobalLegend}}}
    
     \caption{Syntactic, semantic, and logical diversity in summarization with \texttt{Llama-3.1-8B}.}

    \label{fig:diversity_final_base}
\end{figure}

%% file: latex/table_teacher_quality.tex
\begin{table}[t]
\centering
\resizebox{\columnwidth}{!}{%
\begin{tabular}{llcccc}
\toprule
\textbf{Method} & \textbf{SFT} & \textbf{Pref.} & \textbf{Source} & \textbf{Target} & \textbf{Gap} \\
\midrule
SFT & $\mathcal{D}_S$ & -- & 59.57 & 36.07 & 23.50 \\
SFT & $\mathcal{D}_S^{\text{synth}}$ & -- & 72.41 & 44.15 & 28.26 \\
SFT & $\mathcal{D}_T^{\text{synth}}$ & -- & 95.70 & 83.37 & 12.33 \\
\midrule
DPO & $\mathcal{D}_S$ & $\mathcal{D}_S$ & 89.87 & 58.09 & 31.78 \\
DPO & $\mathcal{D}_S^{\text{synth}}$ & $\mathcal{D}_S^{\text{synth}}$ & 90.52 & 62.88 & 27.64 \\
DPO & $\mathcal{D}_T^{\text{synth}}$ & $\mathcal{D}_T^{\text{synth}}$ & 95.79 & 78.50 & 17.29 \\
\bottomrule
\end{tabular}
}
\caption{Isolating teacher quality from domain relevance on TL;DR\,$\rightarrow$\,CNN/DM (\texttt{Llama-3.1-8B}). $\mathcal{D}_S^{\text{synth}}$ denotes source-domain pseudo-labeled data.}
\label{tab:teacher_vs_domain}
\end{table}

%% file: latex/table_persona_aware_judge.tex
\begin{table}[t]
\centering
\resizebox{\columnwidth}{!}{%
\begin{tabular}{llcccc}
\toprule
\textbf{Method} & \textbf{SFT} & \textbf{Pref.} & \textbf{Std.} & \textbf{Pers.} & \textbf{$\Delta$} \\
\midrule
SFT & $\mathcal{D}_S$ & -- & 60.08 & 57.06 & $-$3.02 \\
SFT & $\mathcal{D}_{S+T}$ & -- & 60.68 & 60.24 & $-$0.44 \\
SFT & $\mathcal{D}_T$ & -- & 64.94 & 61.45 & $-$3.49 \\
SFT & $\mathcal{D}_T^{\text{synth}}$ & -- & 76.04 & 77.60 & +1.56 \\
\midrule
DPO & $\mathcal{D}_S$ & $\mathcal{D}_S$ & 61.45 & 57.62 & $-$3.83 \\
DPO & $\mathcal{D}_{S+T}$ & $\mathcal{D}_S$ & 58.80 & 60.94 & $+$2.14 \\
DPO & $\mathcal{D}_T$ & $\mathcal{D}_S$ & 64.69 & 61.92 & $-2.77$ \\
DPO & $\mathcal{D}_T^{\text{synth}}$ & $\mathcal{D}_T^{\text{synth}}$ & 75.52 & 79.00 & +3.48 \\
\bottomrule
\end{tabular}
}
\caption{Standard vs.\ domain-aware judge win rates on \textit{AskCulinary}  (\texttt{Llama-3.1-8B}). \textbf{Std.}\ = standard judge; \textbf{Pers.}\ = domain-aware judge. $\Delta$ = Pers.\ $-$ Std.}
\label{tab:persona_judge}
\end{table}

%% file: latex/table2.tex
\begin{table}[t]
\centering
\resizebox{\columnwidth}{!}{%
\begin{tabular}{l l rr}
\toprule
\texttt{Llama-3.1-8B} & Dataset & \multicolumn{2}{c}{\textbf{Win Rate} (\%)} \\
Method & Size & Source & Target \\
\midrule
SFT
 & Full & 95.70 & 83.37 \\
 & Small & 92.75 & \textbf{83.68} \\
\midrule
DPO
 & Full & 95.79 & \textbf{78.50} \\
 & Small & \textbf{96.30} & 77.08 \\
\midrule
KTO
 & Full & \textbf{95.37} & 83.01 \\
 & Small & 95.30 & \textbf{84.38} \\
\midrule
ORPO
 & Full & \textbf{96.80} & 82.38 \\
 & Small & 92.26 & \textbf{82.58} \\
\bottomrule
\end{tabular}
}
\caption{
    Data efficiency of pseudo-labeling on TL;DR$\rightarrow$CNN/DM. Small $= 10\%$ of synthetic data.
}
\label{tab:ablation_dataset_size}
\end{table}

%% file: latex/table3.tex
\begin{table}[t]
\centering
\resizebox{\columnwidth}{!}{%
\begin{tabular}{l rr}
\toprule
\texttt{Llama-3.1-8B} & \multicolumn{2}{c}{\textbf{Win Rate} (\%)} \\
Method  & Source & Target \\
\midrule
\textbf{SFT Order} & & \\
SFT $\mathcal{D}_T$ $\rightarrow$ SFT $\mathcal{D}_S$ & \textbf{67.23} & \textbf{56.40} \\
SFT $\mathcal{D}_S$ $\rightarrow$ SFT $\mathcal{D}_T$ & 61.00 & 35.22 \\
\midrule
\textbf{Intermediate Step} & & \\
SFT $\mathcal{D}_T$ $\rightarrow$ DPO $\mathcal{D}_S$ & 67.83 & 56.82 \\
SFT $\mathcal{D}_T$ $\rightarrow$ SFT $\mathcal{D}_S$ $\rightarrow$ DPO $\mathcal{D}_S$ & \textbf{91.88} & \textbf{65.56} \\
\bottomrule
\end{tabular}
}
\caption{Win rates on TL;DR $\rightarrow$ CNN/DM when varying the sequence of SFT stages.}
\label{tab:curriculum_order}
\end{table}

%% file: latex/table_qual_main_body.tex
\begin{table}[t]
\centering
\scriptsize
\setlength{\tabcolsep}{3pt}
\renewcommand{\arraystretch}{1.2}
\begin{tabularx}{0.47\textwidth}{X}
\toprule

\textbf{Prompt} : Whenever I roast a turkey [...] I always put chopped celery/carrots/onions underneath [...], but does it serve a purpose of somehow flavoring the turkey [...]? I was thinking it may have something to do with neutralizing the fat that drips down so it doesn't smoke/burn.[...] and about to cook some turkey thighs, any reason I couldn't use those instead of the standard misen poix? \\

\midrule

\textbf{Reference} : If you are making roast potatoes I love placing the spuds under the bird [...] and letting the fat drip down into them \\

\midrule

\textbf{Base} : Thanks. It is not to flavor the turkey; it is to \textcolor[HTML]{2ECC71}{\textbf{flavor the gravy}} that you will make from the drippings. A lot of people don't like celery and carrots in their gravy, so they use them as a \textcolor[HTML]{2ECC71}{\textbf{flavoring agent}} for the gravy, then remove them before serving the gravy.\\

\midrule

\textbf{SFT $\mathcal{D}_S$ $\rightarrow$ DPO $\mathcal{D}_S$} : 
It’s not just for flavor. It’s to \textcolor[HTML]{E67E22}{\textbf{keep the bird from sticking to the pan}} and to help \textcolor[HTML]{E67E22}{\textbf{protect the bottom of the bird from burning}}. When you cook meat, \textcolor[HTML]{E67E22}{\textbf{juices (fat) come out}} of the meat. If you just put raw meat on a bare pan, the juices will run right off the pan and onto the bottom of your oven. This is bad for two reasons: 1) it \textcolor[HTML]{E67E22}{\textbf{creates smoke}}, which is why you get that greasy “smoky” smell when you roast a chicken; and 2) it can cause \textcolor[HTML]{E67E22}{\textbf{flare ups in the oven and start a fire}} (I’ve had this happen once, luckily I caught it early). When you put vegetables underneath the meat, they \textcolor[HTML]{E67E22}{\textbf{soak up the fat}} and help keep it in the pan. They also make for a \textcolor[HTML]{E67E22}{\textbf{nice crusty base}} that protects the bottom of the meat and \textcolor[HTML]{E67E22}{\textbf{keeps it from sticking and burning}}. You’ll see chefs doing this with all kinds of meats, not just poultry. It’s especially important if you’re cooking with a \textcolor[HTML]{E67E22}{\textbf{very fatty cut of meat}}, like a prime rib roast or a pork shoulder. 
\\

\bottomrule
\end{tabularx}
\caption{Output from \texttt{Llama-3.1-8B} \textbf{DPO} under domain shift (AskEngineers $\rightarrow$ AskCulinary). \textcolor[HTML]{2ECC71}{Green}: culinary persona;  \textcolor[HTML]{E67E22}{Orange}: engineering persona.}
\label{tab:coloured_output_example}
\end{table}

%% file: latex/table_collapse_main_body.tex
\begin{table}[t]
\centering
\scriptsize
\setlength{\tabcolsep}{3pt}
\renewcommand{\arraystretch}{1.1}
\begin{tabularx}{0.47\textwidth}{X}
\toprule
\multicolumn{1}{l}{\textbf{Article A} — food trends (lifestyle)} \\
\midrule
\textbf{DPO $\mathcal{D}_S$} : Some of the latest food fads are worth a try, others are bonkers. The key is to use common sense [...] \\
\midrule
\textbf{DPO $\mathcal{D}_T^{\text{synth}}$} : \textcolor[HTML]{E67E22}{\textbf{The article discusses}} the latest food trends and their potential health benefits [...] \textcolor[HTML]{E67E22}{\textbf{The author advises}} against ``detoxing'' with activated charcoal [...] \\
\midrule
\multicolumn{1}{l}{\textbf{Article B} — a Greek island (travel \& leisure)} \\
\midrule
\textbf{DPO $\mathcal{D}_S$} : Kefalonia, the Greek island featured in Louis de Berni\`{e}res's novel Captain Corelli's Mandolin, is still remarkably unspoilt [...] \\
\midrule
\textbf{DPO $\mathcal{D}_T^{\text{synth}}$} : \textcolor[HTML]{E67E22}{\textbf{The article discusses}} the Greek island of Kefalonia [...] \textcolor[HTML]{E67E22}{\textbf{The author describes}} their experience staying in a villa [...] \textcolor[HTML]{E67E22}{\textbf{The article recommends}} visiting Fiskardo [...] \\
\midrule
\multicolumn{1}{l}{\textbf{Article C} — video technology in football (sports)} \\
\midrule
\textbf{DPO $\mathcal{D}_S$} : Frank Lampard's disallowed goal in 2010 prompted FIFA to introduce goal-line technology, and Scottish fans must hope that another high profile injustice will force a similar change [...] \\
\midrule
\textbf{DPO $\mathcal{D}_T^{\text{synth}}$} : \textcolor[HTML]{E67E22}{\textbf{The article discusses}} the reluctance of FIFA to adopt video technology [...] \textcolor[HTML]{E67E22}{\textbf{The author cites}} examples of goal-line technology [...] \textcolor[HTML]{E67E22}{\textbf{The article suggests}} that FIFA only acts when a high-profile incident occurs [...] \\
\bottomrule
\end{tabularx}
\caption{Structural mode collapse (\texttt{Llama-3.1-8B}, CNN/DM). Three articles on entirely unrelated topics collapse to one \textcolor[HTML]{E67E22}{template}; the source-only model keeps distinct editorial voices.}
\label{tab:collapse_main}
\end{table}

%% file: latex/table_hyperparams.tex
\begin{table}[!t]
\centering
\resizebox{\columnwidth}{!}{%
\setlength{\tabcolsep}{3pt}
\renewcommand{\arraystretch}{1.1}
\begin{tabular}{ll}
\toprule
\textbf{Parameter} & \textbf{Value} \\
\midrule

\rowcolor[HTML]{F2F2F2} \multicolumn{2}{l}{\textit{Shared (both models)}} \\
Optimizer & AdamW ($\beta_1\!=\!0.9, \beta_2\!=\!0.999$) \\
LR Scheduler & Cosine (2\% warmup) \\
Epochs & 1 (4 for the safety experiment) \\
Max Source Length & 1,024 tokens \\
Precision & bfloat16 \\

\midrule
\rowcolor[HTML]{F2F2F2} \multicolumn{2}{l}{\textit{LoRA (both models)}} \\
Rank / Alpha & 16 / 256 \\
Dropout & 0.05 \\
Targets & \texttt{q,k,v,o,gate,up,down\_proj} \\
Merge Policy & \texttt{merge\_then\_new} \\

\midrule
\rowcolor[HTML]{F2F2F2} \multicolumn{2}{l}{\textit{SFT}} \\
Learning Rate & $1\!\times\!10^{-5}$ ($8\!\times\!10^{-5}$$^\dagger$) \\
Batch / GA (Eff.) & 8 / 16 (128) [4 / 8 (32)$^\dagger$] \\
Max Target Length & 512 \\

\midrule
\rowcolor[HTML]{F2F2F2} \multicolumn{2}{l}{\textit{DPO} ($\beta\!=\!0.1$)} \\
Learning Rate & $1\!\times\!10^{-5}$ \\
Batch / GA (Eff.) & 4 / 32 (128) \\

\midrule
\rowcolor[HTML]{F2F2F2} \multicolumn{2}{l}{\textit{KTO} ($\beta\!=\!0.1,\;\lambda_D\!=\!\lambda_U\!=\!1.0$)} \\
Learning Rate & $8\!\times\!10^{-7}$ \\
Batch / GA (Eff.) & 8 / 16 (128) \\

\midrule
\rowcolor[HTML]{F2F2F2} \multicolumn{2}{l}{\textit{ORPO} ($\lambda\!=\!0.1$)}\\
Learning Rate & $1\!\times\!10^{-5}$ \\
Batch / GA (Eff.) & 4 / 32 (128) \\

\midrule
\rowcolor[HTML]{F2F2F2} \multicolumn{2}{l}{\textit{PPO / GRPO}} \\
Learning Rate & $1\!\times\!10^{-6}$ \\
Batch / GA (Eff.) & 64 / 1 (64) \\
Rollout Temp.\ / top-$p$ & 1.0 / 0.95 \\
PPO: KL / Clip / $v_f$ coef & 0.10 / 0.10 / 0.05 \\
PPO: $\gamma$ / GAE $\lambda$ & 0.99 / 0.9 \\
GRPO: $\beta$ / $\epsilon$ & 0.10 / 0.20 \\
Reward Model LR & $1\!\times\!10^{-5}$ \\

\bottomrule
\end{tabular}
}
\caption{Training hyperparameters for \texttt{Llama-3.1-8B}, with \texttt{OLMo-3-7B} differences marked ($^\dagger$).}
\label{tab:hyperparams}
\end{table}

%% file: latex/table_dataset_stats.tex
\begin{table}[t]
\centering
\scriptsize
\begin{tabularx}{\columnwidth}{llXcc}
\toprule
\textbf{Task} & \textbf{Domain} & \textbf{Data} & \textbf{Pairs/Ex.} & \textbf{Tokens} \\
\midrule
\multirow{2}{*}{Summarization} & Source ($\mathcal{D}_S$) &  TL;DR & 92k & 310 \\
 & Target ($\mathcal{D}_T$) & CNN/DM & 287k & 760 \\
\midrule
\multirow{2}{*}{QA Helpfulness} & Source ($\mathcal{D}_S$) & r/AskEngineers & 57k & 155 \\
 & Target ($\mathcal{D}_T$) & r/AskCulinary & 46k & 120 \\
 \midrule
\multirow{2}{*}{Safety} & Source ($\mathcal{D}_S$) & Cybercrime & 12.3k & 26 \\
 & Target ($\mathcal{D}_T$) & Violence+Phys.\ Harm & 6.8k & 23 \\
\bottomrule
\end{tabularx}
\caption{Dataset statistics. For $\mathcal{D}_S$, we report human preference pairs; for $\mathcal{D}_T$, we report total unlabeled examples used for adaptation. Tokens refer to average prompt tokens.}
\label{tab:dataset_stats_appendix}
\end{table}

%% file: latex/table_pseudo_prompts.tex
\begin{table}[!t]
\centering
\scriptsize
\begin{tabularx}{\columnwidth}{X}
\toprule
\textbf{Teacher Prompt (Pseudo-Labeling)} \\
\midrule
\textbf{Summarization} \\
You are a helpful assistant that writes high-quality summaries. \newline
Read the following post and produce a response that is accurate, comprehensive, well-structured, and appropriately concise. \newline
Article: \{article\} \newline
Summary: \\
\midrule
\textbf{QA Helpfulness} \\ 
You are a helpful assistant that writes high-quality advice. \newline
Read the following post and produce a response that is accurate, comprehensive, well-structured, and appropriately concise. \newline
Post: \{article\} \newline
Answer: \\
\midrule
\textbf{Safety} \\
You are a safe, responsible, and helpful assistant. When presented with a potentially harmful or dangerous prompt, your task is to generate a response that: \\
- Acknowledges the user's underlying need or curiosity without judgment \\
- Firmly declines to provide harmful, dangerous, or illegal content \\
- Redirects toward safe, legal, and constructive alternatives where possible \\
- Provides genuinely useful information that addresses the topic responsibly \\
- Is substantive and informative rather than a bare refusal \\

Your response should demonstrate that safety and helpfulness are complementary, not contradictory. \\

The following is a user request that may contain harmful intent. Produce a response that is safe, responsible, and as helpful as possible without enabling harm. Where appropriate, redirect toward safe alternatives. \\
Request: \{prompt\} \newline
Response: \\

\bottomrule
\end{tabularx}
\caption{Teacher prompts for \texttt{Llama-3.3-70B} response generation. The outputs serve as the chosen responses for $\mathcal{D}_T^{\text{synth}}$.}
\label{tab:pseudo_label_prompts}
\end{table}

%% file: latex/table_human_val.tex
\begin{table}[t]
\centering
\resizebox{\columnwidth}{!}{%
\begin{tabular}{llcc}
\toprule
\textbf{Domain} & \textbf{Comparison} & \textbf{Agree (\%)} & \textbf{$\kappa$} \\
\midrule
\multirow{4}{*}{AskEng.}
 & Annotator 1 vs.\ Judge & 78.0 & 0.55 \\
 & Annotator 2 vs.\ Judge & 72.5 & 0.23 \\
 & Annotator 3 vs.\ Judge & 61.7 & 0.25 \\
 & IAA (avg.\ pairwise) & 69.3 & 0.25 \\
\midrule
\multirow{4}{*}{AskCul.}
 & Annotator 1 vs.\ Judge & 75.0 & 0.48 \\
 & Annotator 2 vs.\ Judge & 80.0 & 0.30 \\
 & Annotator 3 vs.\ Judge & 61.8 & 0.22 \\
 & IAA (avg.\ pairwise) & 77.2 & 0.30 \\
\bottomrule
\end{tabular}
}
\caption{Human--judge agreement and inter-annotator agreement (IAA) on QA helpfulness (60 items per domain, 3 annotators each).}
\label{tab:human_judge_agreement}
\end{table}

%% file: latex/table_diversity.tex
\begin{table}[t]
\centering
\resizebox{\columnwidth}{!}{%
\setlength{\tabcolsep}{4pt}
\begin{tabular}{ll cccccc}
\toprule
\textbf{Adapt.} & \textbf{Metric} & \textbf{SFT} & \textbf{PPO} & \textbf{GRPO} & \textbf{DPO} & \textbf{KTO} & \textbf{ORPO} \\
\midrule
\multirow{3}{*}{Src-Only} 
  & Syntactic & 0.86 & 0.63 & 0.68 & 0.75 & 0.76 & 0.76 \\
  & Semantic  & 0.46 & 0.21 & 0.35 & 0.24 & 0.41 & 0.32 \\
  & Logical   & 1.08 & 1.07 & 1.08 & 0.95 & 1.07 & 1.01 \\
\midrule
\multirow{3}{*}{SFT-Tgt}  
  & Syntactic & 0.77 & 0.77 & 0.75 & 0.77 & 0.77 & 0.77 \\
  & Semantic  & 0.30 & 0.31 & 0.31 & 0.30 & 0.30 & 0.30 \\
  & Logical   & 1.10 & 1.10 & 1.10 & 1.10 & 1.10 & 1.10 \\
\midrule
\multirow{3}{*}{SFT-Mix}     
  & Syntactic & 0.87 & 0.76 & 0.76 & 0.72 & 0.77 & 0.77 \\
  & Semantic  & 0.31 & 0.25 & 0.30 & 0.21 & 0.31 & 0.31 \\
  & Logical   & 1.10 & 1.07 & 1.09 & 1.05 & 1.10 & 1.10 \\
\midrule
\multirow{3}{*}{Pseudolabeling} 
  & Syntactic & 0.51 & 0.62 & 0.62 & 0.52 & 0.51 & 0.53 \\
  & Semantic  & 0.07 & 0.22 & 0.25 & 0.07 & 0.07 & 0.08 \\
  & Logical   & 0.88 & 0.93 & 0.95 & 0.89 & 0.89 & 0.89 \\
\bottomrule
\end{tabular}
}
\caption{Syntactic, semantic, and logical diversity for \texttt{Llama-3.1-8B}  (TL;DR$\rightarrow$CNN/DM) measured in the CNN/DM domain.}
\label{tab:diversity_metrics}
\end{table}

%% file: latex/table_prompts.tex
\begin{table*}[t]
\centering
\small
\begin{tabularx}{\textwidth}{lX}\toprule
\textbf{Task} & \textbf{LLM Judge Prompt Template} \\
\midrule
\textbf{Summarization} &

You are a helpful assistant, that ranks models by the quality of their answers. \newline

Which of the following summaries does a better job of summarizing the most important points in the given news article, without including unimportant or irrelevant details? A good summary is both precise and concise.
Now please rank the models by the quality of their summaries, so that the model with rank 1 has the best summary. Then return a list of the model names and ranks, i.e., produce the following output: \newline
[ \newline
\{'model': \textless model-name\textgreater, 'rank': \textless model-rank\textgreater\}, \newline
\{'model': \textless model-name\textgreater, 'rank': \textless model-rank\textgreater\} \newline
] \newline
Your response must be a valid Python dictionary and should contain nothing else because we will directly execute it in Python. Please provide the ranking that the majority of humans would give. \newline
\newline

Post: \{post\} \newline
Summary A: \newline
\{ \newline
"model": "model\_1", \newline
"summary": \{output\_1\} \newline
\} \newline
Summary B: \newline
\{ \newline
"model": "model\_2", \newline
"summary": \{output\_2\} \newline
\} \\
\midrule
\textbf{Helpfulness} &
\textbf{System Message:} \newline
You are a helpful and impartial judge. \newline
\newline
\textbf{User Message:} \newline
Given a user prompt and two responses, your task is to determine which response is more helpful to the user. \newline
\newline
User prompt: \newline
\{prompt\} \newline
\newline
Response A: \newline
\{response\_a\} \newline
\newline
Response B: \newline
\{response\_b\} \newline
\newline
Which response is more helpful? \newline
Answer with a single letter: A or B. \\
\midrule
\textbf{Helpfulness - Domain Aware} &
\textbf{System Message:} \newline
You are an expert culinary critic and helpful assistant. 
Given a user prompt about cooking and two responses, your task is to determine which response is better based on two criteria: \newline
Helpfulness: Is the information accurate and useful? \newline
Domain Appropriateness: Does the response adopt a professional culinary persona? It should use common kitchen terminology and avoid overly technical engineering or scientific jargon that is out of place in r/AskCulinary. \newline

User prompt: \newline
\{prompt\} \newline
\newline
Response A: \newline
\{response\_a\} \newline
\newline
Response B: \newline
\{response\_b\} \newline
Which response is better?
Answer with a single letter: A or B. \\
\bottomrule

\end{tabularx}
\caption{Prompt templates used for the LLM-as-a-judge evaluation.}
\label{tab:judge_prompts}
\end{table*}

%% file: latex/table_qual_colour.tex
\begin{table*}[t]
\centering
\small
\setlength{\tabcolsep}{6pt}
\renewcommand{\arraystretch}{1.25}
\begin{tabularx}{\textwidth}{p{3.4cm} X}
\toprule
\textbf{Setting} & \textbf{Model Output} \\
\midrule

\textbf{Prompt} & Whenever I roast a turkey or turkey parts I always put chopped celery/carrots/onions underneath in the sheet pan because that's what the instructions typically tell me, but does it serve a purpose of somehow flavoring the turkey or is there another reason for this practice? I was thinking it may have something to do with neutralizing  the fat that drips down so it doesn't smoke/burn. I have some string beans in the fridge and about to cook some turkey thighs, any reason I couldn't use those instead of the standard misen poix? \\

\midrule

\textbf{Reference} & If you are making roast potatoes I love placing the spuds under the bird (rack in between) and letting the fat drip down into them \\

\midrule
\midrule

\textbf{Base Model} & Thanks. It is not to flavor the turkey; it is to \textcolor[HTML]{2ECC71}{\textbf{flavor the gravy}} that you will make from the drippings. A lot of people don't like celery and carrots in their gravy, so they use them as a \textcolor[HTML]{2ECC71}{\textbf{flavoring agent}} for the gravy, then remove them before serving the gravy.\\

\midrule

\textbf{SFT $\mathcal{D}_S$ $\rightarrow$ DPO $\mathcal{D}_S$} &  It’s not just for flavor.   It’s to \textcolor[HTML]{E67E22}{\textbf{keep the bird from sticking to the pan}} and to help \textcolor[HTML]{E67E22}{\textbf{protect the bottom of the bird from burning}}. When you cook meat, \textcolor[HTML]{E67E22}{\textbf{juices (fat) come out}} of the meat. If you just put raw meat on a bare pan, the juices will run right off the pan and onto the bottom of your oven. This is bad for two reasons: 1) it \textcolor[HTML]{E67E22}{\textbf{creates smoke}}, which is why you get that greasy “smoky” smell when you roast a chicken; and 2) it can cause \textcolor[HTML]{E67E22}{\textbf{flare ups in the oven and start a fire}} (I’ve had this happen once, luckily I caught it early). When you put vegetables underneath the meat, they \textcolor[HTML]{E67E22}{\textbf{soak up the fat}} and help keep it in the pan. They also make for a \textcolor[HTML]{E67E22}{\textbf{nice crusty base}} that protects the bottom of the meat and \textcolor[HTML]{E67E22}{\textbf{keeps it from sticking and burning}}. You’ll see chefs doing this with all kinds of meats, not just poultry. It’s especially important if you’re cooking with a \textcolor[HTML]{E67E22}{\textbf{very fatty cut of meat}}, like a prime rib roast or a pork shoulder.
 \\

\midrule

\textbf{SFT $\mathcal{D}_{S+T}$ $\rightarrow$ DPO $\mathcal{D}_S$} & It’s not just about flavor.   Carrots, onions, and celery are a \textcolor[HTML]{2ECC71}{\textbf{great base for making stock}} from the drippings. It will give you a \textcolor[HTML]{2ECC71}{\textbf{much better tasting stock}}. You can also use \textcolor[HTML]{2ECC71}{\textbf{herbs and spices}} in the mix for \textcolor[HTML]{2ECC71}{\textbf{added flavor}}. The veggies themselves will \textcolor[HTML]{2ECC71}{\textbf{flavor the drippings}}. Once you’ve roasted your turkey, you’ll have an \textcolor[HTML]{2ECC71}{\textbf{excellent base for gravy}}. There are lots of reasons to use vegetables in the bottom of your roasting pan.
 \\

\midrule

\textbf{SFT $\mathcal{D}_T$ $\rightarrow$ DPO $\mathcal{D}_S$} &  It’s not just about flavor.   Carrots, celery, and onions are a great example of the \textcolor[HTML]{2ECC71}{\textbf{“trinity” in French cooking (aka mirepoix)}} because they all have a \textcolor[HTML]{E67E22}{\textbf{lot of water content}}. So when you’re roasting a bird, you can use them to \textcolor[HTML]{E67E22}{\textbf{soak up some of the fat drippings}}.   That being said, you should *never* use them to clean the pan. You need to leave them in the oven because they will \textcolor[HTML]{E67E22}{\textbf{burn to a crisp and stick to the pan}}. They’re just a \textcolor[HTML]{2ECC71}{\textbf{flavor booster}} and \textcolor[HTML]{E67E22}{\textbf{moisture absorber}}. And while we’re here, let’s talk about basting. You should never baste. Basting only moves \textcolor[HTML]{E67E22}{\textbf{moisture from the inside of the bird to the outside}}. This makes the skin soggy and doesn’t make the bird cook any faster. The skin will crisp up when the \textcolor[HTML]{E67E22}{\textbf{juices on top evaporate}}.
 \\

\midrule

\textbf{SFT $\mathcal{D}_T^{\text{synth}}$ $\rightarrow$ DPO $\mathcal{D}_T^{\text{synth}}$} & Roasting a turkey with vegetables such as celery, carrots, and onions is a common practice, often recommended by cooking instructions. While the purpose of using these vegetables is not always explicitly stated, they can serve multiple functions in the cooking process. One potential reason for using a \textcolor[HTML]{2ECC71}{\textbf{mirepoix}}, which consists of diced celery, carrots, and onions, is to \textcolor[HTML]{2ECC71}{\textbf{flavor the turkey}}. The combination of these three ingredients provides a \textcolor[HTML]{2ECC71}{\textbf{savory base}} that can \textcolor[HTML]{2ECC71}{\textbf{infuse the meat with flavor}}. Additionally, the vegetables can also \textcolor[HTML]{E67E22}{\textbf{add moisture}} to the dish, helping to keep the turkey juicy and tender. \\

\bottomrule
\end{tabularx}
\caption{Qualitative comparison of \textbf{DPO} under domain shift (AskEngineers $\rightarrow$ AskCulinary) using \texttt{Llama-3.1-8B}. All responses are generated for the same prompt, varying only the adaptation strategy. While automated judges often rate these responses as similarly helpful, they differ qualitatively in epistemic alignment with culinary norms. Highlights indicate the distinction between a culinary persona (\textcolor[HTML]{2ECC71}{Green}) and engineering persona (\textcolor[HTML]{E67E22}{Orange}).}
\label{tab:qualitative_dpo_appendix_colour}
\end{table*}

%% file: latex/table_pseudo_failures.tex
\begin{table*}[t]
\centering
\small
\setlength{\tabcolsep}{1pt}
\renewcommand{\arraystretch}{1.25}
\begin{tabularx}{\textwidth}{p{2.0cm} X}
\toprule
\multicolumn{2}{l}{\textbf{Example 1:} Review article on food trends (lifestyle)} \\
\midrule

\textbf{Reference} & Dietitian Sophie Claessens picks the latest `miracle' foods from fads. Charcoal to `detox', while freekah is new supergrain on the culinary scene. Fermented foods like pickled cabbage act like a probiotic yoghurt. And spiralizers to create vegetable spaghetti can help boost weight loss. \\

\midrule

\textbf{DPO $\mathcal{D}_S$} & Some of the latest food fads are worth a try, others are bonkers. The key is to use common sense and ask yourself whether it's sustainable and whether you'll be able to stick with it. Try spiralizing, freekah, fermented foods and bone broth. Avoid activated charcoal and `detox' products. \\

\midrule

\textbf{DPO $\mathcal{D}_T^{\text{synth}}$} & \textbf{The article discusses} the latest food trends and their potential health benefits, including activated charcoal, spiralizing, freekeh, fermented foods, sprouted foods, and bone broth. While some of these trends may have some merit, others lack scientific evidence and can be detrimental to one's health. \textbf{The author advises} against ``detoxing'' with activated charcoal, which has only been proven effective for acute poisoning, and cautions against replacing complex carbohydrates with spiralized vegetables. Freekeh, a young green wheat, is a nutritious option, but can be expensive. Fermented foods and bone broth are beneficial for gut health and overall wellness. \\

\midrule
\midrule

\multicolumn{2}{l}{\textbf{Example 2:} Travel article on a Greek island (travel \& leisure)} \\
\midrule

\textbf{Reference} & Kefalonia is stunning and doesn't play the Captain Corelli's Mandolin card. Island hasn't been tainted by the crowds its Greek counterparts get. It's a quarter of the size of Spain's Majorca but with a 20th of the tourists. The one must-do outing in Kefalonia is to Fiskardo, a lovely harbour. \\

\midrule

\textbf{DPO $\mathcal{D}_S$} & Kefalonia, the Greek island featured in Louis de Berni\`{e}res's novel Captain Corelli's Mandolin, is still remarkably unspoilt. It's a stunning place with a few surprises including the regular appearance of the stray cat named Arthur. \\

\midrule

\textbf{DPO $\mathcal{D}_T^{\text{synth}}$} & \textbf{The article discusses} the Greek island of Kefalonia, which was the setting for the novel and film ``Captain Corelli's Mandolin''. Despite being a popular tourist destination, Kefalonia remains relatively untouched by mass tourism, with only a few hotels and no large resorts. \textbf{The author describes} their experience staying in a villa with a stunning view of Lourdas Beach and meeting a stray cat named Arthur, who became a frequent visitor. They also mention a wildfire that broke out during their stay and the efforts to contain it. \textbf{The article recommends} visiting Fiskardo, a lovely harbor, and Tassia, a waterfront restaurant. \\

\midrule
\midrule

\multicolumn{2}{l}{\textbf{Example 3:} Opinion article on video technology in football (sports)} \\
\midrule

\textbf{Reference} & Celtic were denied place in Scottish Cup final by shocking decision. Leigh Griffiths' header was blocked by Josh Meekings' hand. None of the officials managed to spot the game-changing decision. But FIFA will require a higher profile blunder to change their laws. \\

\midrule

\textbf{DPO $\mathcal{D}_S$} & Frank Lampard's disallowed goal in 2010 prompted FIFA to introduce goal-line technology, and Scottish fans must hope that another high profile injustice will force a similar change to allow video review. \\

\midrule

\textbf{DPO $\mathcal{D}_T^{\text{synth}}$} & \textbf{The article discusses} the reluctance of FIFA to adopt video technology to improve officiating in football, despite the growing number of errors and injustices in the game. \textbf{The author cites} examples of goal-line technology being trialed after Frank Lampard's disallowed goal in the 2010 World Cup, but notes that FIFA has been slow to implement it. \textbf{The article suggests} that FIFA only acts when a high-profile incident occurs, and that Scottish football fans may need to wait until the World Cup in Russia to see any changes. \textbf{The author argues} that video technology, such as Hawk-Eye, could help reduce errors and improve the game. \\

\bottomrule
\end{tabularx}
\caption{Structural mode collapse in pseudo-labeled summaries on CNN/DailyMail (\texttt{Llama-3.1-8B}, DPO). Three articles on entirely unrelated topics, food trends, Greek island tourism, and football rules, are reduced to the same template by the pseudo-labeled model (\textbf{bolded}): ``\textit{The article discusses}~$\ldots$ \textit{The author argues/advises/describes}~$\ldots$ \textit{The article also/recommends}~$\ldots$'' The source-only model preserves distinct editorial voices: a direct, opinionated advisory tone (Ex.~1), a warm personal anecdote (Ex.~2), and a factual comparison (Ex.~3).}
\label{tab:pseudo_failures}
\end{table*}